\documentclass{article}
\usepackage{multirow}
\usepackage{algorithm, algorithmic}
\usepackage{amsthm, amsmath, amssymb, mathtools, multicol}
\usepackage{wrapfig}
\usepackage{xcolor}
\usepackage{natbib}
\usepackage{hyperref}
\usepackage{fancyhdr, fullpage}
\usepackage{graphicx, array, subcaption}
\usepackage[american]{babel}
\usepackage{microtype}
\usepackage{enumitem}
\usepackage{cases}
\usepackage{etoolbox}
\usepackage{siunitx}

\robustify{\fbox}

\newcommand{\R}{\mathbb{R}}
\newcommand{\mE}{\mathbb{E}}
\newcommand{\cond}{\,\middle\vert\,}
\newcommand{\pf}[1]{\nabla f (#1)}

\newcommand{\ph}[1]{\nabla h(#1)}
\newcommand{\phii}[1]{\nabla h_i(#1)}
\newcommand{\phik}[1]{\nabla h_{i_k}(#1)}
\newcommand{\pfik}[1]{\nabla f_{i_k}(#1)}
\newcommand{\pfi}[1]{\nabla f_{i}(#1)}
\newcommand{\pfr}[2]{\nabla f_{#1}(#2)}

\newcommand{\norm}[1]{\left\lVert#1\right\rVert}
\newcommand{\norml}[1]{\left\lVert#1\right\rVert}
\newcommand{\normb}[1]{\big\lVert#1\big\rVert}

\newcommand{\abs}[1]{\left\lvert#1\right\rvert}

\newcommand{\innr}[1]{\left\langle#1\right\rangle}

\newcommand{\Eik}[1]{\mE_{i_k} \left[#1\right]}

\newcommand{\E}[1]{\mE \big[#1\big]}

\newcommand{\xs}{x^{\star}}

\DeclareMathOperator*{\prox}{prox}
\newcommand{\tprox}{\text{$\prox$}}

\newtheorem{lemmas}{Lemma}
\newtheorem{app-lemmas}{Lemma}

\newtheorem{theorems}{Theorem}
\newtheorem{propositions}{Proposition}[theorems]
\newtheorem{app-theorems}{Theorem}[section]
\newtheorem{remark}{Remark}[theorems]
\newtheorem{lemma-remark}{Remark}

\graphicspath{{./Img/}}

\begin{document}

\title{Boosting First-Order Methods by Shifting Objective:\\ New Schemes with Faster Worst-Case Rates}

\author{Kaiwen Zhou\thanks{Department of Computer Science and Engineering, The Chinese University of Hong Kong, Sha Tin, N.T., Hong Kong SAR; e-mail: \href{mailto:kwzhou@cse.cuhk.edu.hk}{\tt kwzhou@cse.cuhk.edu.hk}.} \and Anthony Man-Cho So\thanks{Department of Systems Engineering and Engineering Management, The Chinese University of Hong Kong, Sha Tin, N.T., Hong Kong SAR; e-mail: \href{mailto:manchoso@se.cuhk.edu.hk}{\tt manchoso@se.cuhk.edu.hk}.} \and James Cheng\thanks{Department of Computer Science and Engineering, The Chinese University of Hong Kong, Sha Tin, N.T., Hong Kong SAR; e-mail: \href{mailto:jcheng@cse.cuhk.edu.hk}{\tt jcheng@cse.cuhk.edu.hk}.}}

\pagestyle{plain}
\date{}

\maketitle

\begin{abstract}
We propose a new methodology to design first-order methods for unconstrained strongly convex problems. Specifically, instead of tackling the original objective directly, we construct a shifted objective function that has the same minimizer as the original objective and encodes both the smoothness and strong convexity of the original objective in an interpolation condition. We then propose an algorithmic template for tackling the shifted objective, which can exploit such a condition. Following this template, we derive several new accelerated schemes for problems that are equipped with various first-order oracles and show that the interpolation condition allows us to vastly simplify and tighten the analysis of the derived methods. In particular, all the derived methods have faster worst-case convergence rates than their existing counterparts. Experiments on machine learning tasks are conducted to evaluate the new methods.
\end{abstract}

\section{Introduction}

In this paper, we focus on the following unconstrained smooth strongly convex problem:
\begin{equation}\label{prob_def}
	\min_{x\in \R^d} {f(x) = \frac{1}{n} \sum_{i=1}^n{f_i(x)}},
\end{equation}
where each $f_i$ is $L$-smooth and $\mu$-strongly convex,\footnote{The formal definitions of smoothness, strong convexity are given in Section \ref{sec:notations}. If each $f_i(\cdot)$ is $L$-smooth, the averaged function $f(\cdot)$ is itself $L$-smooth --- but typically with a smaller $L$. We keep $L$ as the smoothness constant for consistency.} and we denote $\xs\in \R^d$ as the solution of this problem. The $n=1$ case covers a large family of classic strongly convex problems, for which gradient descent (GD) and Nesterov's accelerated gradient (NAG) \citep{AGD1,AGD4,AGD3} are the methods of choice. The $n \geq 1$ case is the popular finite-sum case, where many elegant methods that incorporate the idea of variance reduction have been proposed. Problems with a finite-sum structure arise frequently in machine learning and statistics, such as empirical risk minimization (ERM).  

In this work, we tackle problem \eqref{prob_def} from a new angle. Instead of designing methods to solve the original objective function $f$, we propose methods that are designed to solve a shifted objective $h$:
\[
\min_{x\in \R^d} {h(x) = \frac{1}{n} \sum_{i=1}^n {h_i(x)}}, \text{ where } h_i(x) = f_i(x) - f_i(\xs) - \innr{\pfi{\xs}, x - \xs} - \frac{\mu}{2} \norm{x - \xs}^2.\] It can be easily verified that each $h_i(x)$ is $(L-\mu)$-smooth and convex, $\phii{x} = \pfi{x} - \pfi{\xs} - \mu(x - \xs)$, $\ph{x} = \pf{x} - \mu (x - \xs)$, $h_i(\xs) = h(\xs) = 0$ and $ \phii{\xs} = \ph{\xs} = \mathbf{0}$, which means that the shifted problem and problem \eqref{prob_def} share the same optimal solution $\xs$. Let us write a well-known property of $h$: 
\begin{equation}\label{interpolation}
	\forall x, y \in \R^d, h(x) - h(y) - \innr{\ph{y}, x - y} \geq \frac{1}{2(L - \mu)} \norm{\ph{x} - \ph{y}}^2,
\end{equation}
which encodes both the smoothness and strong convexity of $f$. The discrete version of this inequality is equivalent to the \textit{smooth strongly convex interpolation condition} discovered in \citet{taylor2017smooth}. As studied in \citet{taylor2017smooth}, this type of inequality forms a necessary and sufficient condition for the existence of a smooth strongly convex $f$ interpolating a given set of triples $\{(x_i, \nabla f_i, f_i)\}$, while the usual collection of $L$-smoothness and strong convexity inequalities is only a necessary condition.\footnote{It implies that those inequalities may allow a non-smooth $f$ interpolating the set, and thus a worst-case rate built upon those inequalities may not be achieved by any smooth $f$ (i.e., the rate is loose). See \citet{taylor2017smooth} for details.} For worst-case analysis, it implies that tighter results can be derived by exploiting condition $\eqref{interpolation}$ than using smoothness and strong convexity ``separately'', which is common in existing worst-case analysis. We show that our methodology effectively exploits this condition and consequently,  we propose several methods that achieve faster worst-case convergence rates than their existing counterparts. 

In summary, our methodology and proposed methods have the following distinctive features:
\begin{itemize}
	\item We show that our methodology works for problems equipped with various first-order oracles: deterministic gradient oracle, incremental gradient oracle and incremental proximal point oracle.
	\item We leverage a cleaner version of the interpolation condition discovered in \citet{taylor2017smooth}, which leads to simpler and tighter analysis to the proposed methods than their existing counterparts.
	\item For our proposed stochastic methods, we deal with shifted variance bounds / shifted stochastic gradient norm bounds, which are different from all previous works.
	\item All the proposed methods achieve faster worst-case convergence rates than their counterparts that were designed to solve the original objective $f$.
\end{itemize}


Our work is motivated by a recently proposed robust momentum method \citep{RMM}, which converges under a Lyapunov function that contains a term $h(x) - \frac{1}{2(L-\mu)}\norm{\ph{x}}^2$. 
Our work conducts a comprehensive study of the special structure of this term.

This paper is organized as follows: In Section \ref{sec:lemmas}, we present high-level ideas and lemmas that are the core building blocks of our methodology. In Section \ref{sec:deterministic}, we propose an accelerated method for the $n=1$ case. In Section~\ref{sec:variance-reduction}, we propose accelerated stochastic variance-reduced methods for the $n \geq 1$ case with incremental gradient oracle. In Section \ref{sec:proximal-point}, we propose an accelerated method for the $n \geq 1$ case with incremental proximal point oracle. In Section \ref{sec:evaluations}, we provide experimental results.

\subsection{Notations and Definitions}
\label{sec:notations}
In this paper, we consider problems in the standard Euclidean space denoted by $\mathbb{R}^d$. We use $\innr{\cdot,\cdot}$ and $\norm{\cdot}$ to denote the inner product and the Euclidean norm, respectively. We let $[n]$ denote the set $\{1,2,\ldots,n\}$, $\mE$ denote the total expectation and $\mE_{i_k}$ denote the conditional expectation given the information up to iteration~$k$. 

We say that a convex function $f: \R^d \rightarrow \R$ is \textit{$L$-smooth} if it has $L$-Lipschitz continuous gradients, i.e.,
\[
\forall x, y\in \R^d, \norm{\pf{x} - \pf{y}} \leq L\norm{x - y}.
\]
Some important consequences of this assumption can be found in the textbook \citep{AGD3}: \[\forall x, y\in \R^d,
\frac{1}{2L} \norm{\pf{x} - \pf{y}}^2 \leq f(x) - f(y) - \innr{\pf{y}, x - y} \leq \frac{L}{2}\norm{x - y}^2.
\]
We refer to the first inequality as \textit{interpolation condition} following \citet{taylor2017smooth}. A continuously differentiable $f$ is called \textit{$\mu$-strongly convex} if
\[
\forall x, y\in \R^d, f(x) - f(y) - \innr{\pf{y}, x - y} \geq \frac{\mu}{2} \norm{x - y}^2.
\] 

Given a point $x\in \R^d$, an index $i\in [n]$ and $\alpha > 0$, a deterministic oracle returns $(f(x), \pf{x})$, an incremental first-order oracle returns $(f_i(x), \pfi{x})$ and an incremental proximal point oracle returns $(f_i(x), \pfi{x}, \tprox_i^{\alpha}(x))$, where the proximal operator is defined as $\tprox_i^\alpha (z) = \arg\min_x{\{f_i(x) + \frac{\alpha}{2} \norm{x - z}^2\}}$. We denote $\epsilon > 0$ as the required accuracy for solving problem \eqref{prob_def} (i.e., to achieve $\norm{x - \xs}^2 \leq \epsilon$), which is assumed to be small. We denote $\kappa \triangleq \frac{L}{\mu}$, which is often called the condition ratio.

\subsection{Related Work}

Problem \eqref{prob_def} with $n=1$ is the classic smooth strongly convex setting. Standard analysis shows that for this problem, GD with $\frac{2}{L + \mu}$ stepsize converges linearly at a $(\frac{\kappa - 1}{\kappa + 1})^2$ rate\footnote{In this paper, the worst-case convergence rate is measured in terms of the squared norm distance $\norm{x - \xs}^2$.} (see the textbook \citep{AGD3}). The heavy-ball method \citep{polyak1964some} fails to converge globally on this problem \citep{IQC}. The celebrated NAG is proven to achieve a faster $1 - 1/\sqrt{\kappa}$ rate \citep{AGD3}. This rate remains the fastest one until recently, \citet{TM} proposed the Triple Momentum method (TM) that converges at a $(1 - 1/\sqrt{\kappa})^2$ rate. Numerical results in \citet{lessard2019direct} suggest that this rate is not improvable. In terms of reducing $\norm{x - \xs}^2$ to $\epsilon$, TM is stated to have an $O\big((\sqrt{\kappa}/2) (\log{\frac{1}{\epsilon}} + \log{\sqrt{\kappa}})\big)$ iteration complexity (cf. Table 2, \citep{TM}) compared with the $O(\sqrt{\kappa}\log{\frac{1}{\epsilon}})$ complexity of NAG. 

In the general convex setting, recent works \citep{OGM,attouch2016rate,G-OGM} propose new schemes that have lower complexity than the original NAG. Several of these new schemes were discovered based on the recent works that use semidefinite programming to study worst-case performances of first-order methods. Starting from the performance estimation framework introduced in \citet{PEP}, many different approaches and extensions have been proposed \citep{IQC, computer-aided,taylor2017smooth, taylor2017exact, pmlr-v80-taylor18a}.

For the $n \geq 1$ case, stochastic gradient descent (SGD) \citep{robbins:sgd}, which uses component gradients $\pfi{x}$ to estimate the full gradient $\pf{x}$, achieves a lower iteration cost than GD. However, SGD only converges at a sub-linear rate. To fix this issue, various variance reduction techniques have been proposed recently, such as SAG \citep{SAG,SAG2}, SVRG \citep{SVRG,SVRG2}, SAGA \citep{SAGA}, SDCA \citep{SDCA} and SARAH \citep{SARAH}. Inspired by the Nesterov's acceleration technique, accelerated stochastic variance-reduced methods have been proposed in pursuit of the lower bound $O(n + \sqrt{n\kappa}\log{\frac{1}{\epsilon}})$ \citep{woodworth2016tight}, such as Acc-Prox-SVRG \citep{nitanda:svrg}, APCG \citep{lin:APCG}, ASDCA \citep{ASDCA}, APPA \citep{roy:appa}, Catalyst \citep{lin:vrsg}, SPDC \citep{zhang:spdc}, RPDG \citep{lan2018optimal}, Point-SAGA \citep{point-SAGA} and Katyusha \citep{zhu:Katyusha}. Among these methods, Katyusha and Point-SAGA, representing the first two directly accelerated incremental methods, achieve the fastest rates. Point-SAGA leverages a more powerful incremental proximal operator oracle. Katyusha introduces the idea of negative momentum, which serves as a variance reducer that further reduces the variance of the SVRG estimator. This construction motivates several new accelerated methods \citep{kw:MiG, pmlr-v80-allen-zhu18a,lan2019unified,kulunchakov19a,kw:SSNM, pmlr-v124-zhou20a}.

\section{Tackling the Shifted Objective}\label{sec:lemmas}

As mentioned in the introduction, our methodology is to minimize the shifted objective\footnote{In the Lyapunov analysis framework, this is equivalent to picking a family of Lyapunov function that only involves the shifted objective $h$ (instead of $f$). See \citet{potential} for a nice review of Lyapunov-function-based proofs.} $h$ with the aim of exploiting the interpolation condition. However, a critical issue is that we cannot even compute its gradient $\ph{x}$ (or $\phii{x}$), which requires the knowledge of $\xs$. We figured out that in some simple cases, a change of ``perspective'' is enough to access this gradient information. Take GD $x_{k+1} = x_k - \eta \pf{x_k}$ as an example. Based on the definition $\ph{x_k} = \pf{x_k} - \mu (x_k - \xs)$, we can rewrite the GD update as $x_{k+1} - \xs = (1 - \eta\mu) (x_k - \xs) - \eta \ph{x_k}$, and thus
\[
\norm{x_{k+1} - \xs}^2 = (1 - \eta \mu)^2 \norm{x_k - \xs}^2 \underbrace{- 2\eta(1 - \eta\mu)\innr{\ph{x_k}, x_k - \xs} + \eta^2 \norm{\ph{x_k}}^2}_{R_0}.
\]
If we set $\eta = \frac{2}{L + \mu}$, using the interpolation condition \eqref{interpolation}, we can conclude that $R_0 \leq 0$, which leads to a convergence guarantee. It turns out that this argument is just the one-line proof of GD in the textbook (Theorem 2.1.15, \citep{AGD3}) but looks more structured in our opinion. However, this change of ``perspective'' is too abstract for more complicated schemes. Our solution is to first fix a template updating rule, and then encode this idea into a technical lemma, which serves as an instantiation of the shifted gradient oracle. To facilitate its usage, we formulate this lemma with a classic inequality whose usage has been well-studied. Proofs in this section are given in Appendix \ref{app:lemmas}.

\begin{lemmas}[Shifted mirror descent lemma]\label{p-dis-contract}
	Given a gradient estimator $\mathcal{G}_y$, vectors $z^+, z^-, y\in \R^d$, fix the updating rule $z^+ = \arg\min_{x} \big\{\innr{\mathcal{G}_y, x} + \frac{\alpha}{2} \norm{x - z^-}^2 + \frac{\mu}{2}	\norm{x - y}^2 \big\}$. Suppose that we have a shifted gradient estimator $\mathcal{H}_y$ satisfying the relation $\mathcal{H}_y = \mathcal{G}_y - \mu (y - \xs)$, it holds that
	\[
	\innr{\mathcal{H}_y, z^- - \xs} = \frac{\alpha}{2} \left(\norm{z^- - \xs}^2 - \left(1 + \frac{\mu}{\alpha}\right)^2\norm{z^+ - \xs}^2\right) + \frac{1}{2\alpha} \norm{\mathcal{H}_y}^2.
	\]
\end{lemmas}

\begin{lemma-remark}In general convex optimization, a similar lemma (for $\mathcal{G}$) serves as the core lemma for mirror descent\footnote{In the Euclidean case, mirror descent coincides with GD. It represents a different approach to the same method. } (e.g., Theorem 5.3.1 in the textbook \citep{Aharon2013}). This type of lemma also appears frequently in online optimization, which is used as an upper bound on the regret at the current iteration (e.g., Lemma 3 in \citet{shalev2007logarithmic}). In the strongly convex setting, unlike the common $(1 + \frac{\mu}{\alpha})^{-1}$ (or $1 - \frac{\mu}{\alpha}$) contraction ratio in existing work (e.g., Lemma 2.5 in \citet{zhu:Katyusha}), Lemma \ref{p-dis-contract} provides a $(1 + \frac{\mu}{\alpha})^{-2}$ ratio, which is one of the keys to the improved worst-case rates achieved in this paper.
\end{lemma-remark}

Lemma \ref{p-dis-contract} allows us to choose various gradient estimators for $h$ directly, given that the relation $\mathcal{H}_x = \mathcal{G}_x - \mu (x - \xs)$ holds for some practical $\mathcal{G}_x$. Here we provide some examples:
\begin{itemize}[leftmargin=*]
	\item Deterministic gradient:  $\mathcal{H}_x^{\text{GD}} = \ph{x}\Rightarrow \mathcal{G}_x^{\text{GD}} = \pf{x}.$ 
	\item SVRG estimator:  $\mathcal{H}^{\text{SVRG}}_{x} = \phii{x} - \phii{\tilde{x}} + \ph{\tilde{x}} \Rightarrow \mathcal{G}^{\text{SVRG}}_{x}= \pfi{x} - \pfi{\tilde{x}} + \pf{\tilde{x}}.$
	\item SAGA estimator:  $\mathcal{H}^{\text{SAGA}}_{x} = \phii{x} - \phii{\phi_{i}} + \frac{1}{n}\sum_{j=1}^n{\nabla h_j(\phi_j)} \Rightarrow$\\
	\rule{81.9pt}{0pt}$\mathcal{G}^{\text{SAGA}}_{x} = \pfi{x} - \pfi{\phi_{i}} + \frac{1}{n}\sum_{j=1}^n{\pfr{j}{\phi_j}} - \mu\big(\frac{1}{n} \sum_{j = 1}^n{\phi_j} - \phi_{i}\big).$
	\item SARAH estimator: $\mathcal{H}^{\text{SARAH}}_{x_k} = \phik{x_{k}} - \phik{x_{k-1}} + \mathcal{H}^{\text{SARAH}}_{x_{k-1}}$ and $\mathcal{H}^{\text{SARAH}}_{x_0} = \ph{x_0} \Rightarrow$\\
	\rule{89.2pt}{0pt}$\mathcal{G}^{\text{SARAH}}_{x_k} = \pfik{x_{k}} - \pfik{x_{k-1}} + \mathcal{G}^{\text{SARAH}}_{x_{k-1}}$ and $\mathcal{G}^{\text{SARAH}}_{x_0} = \pf{x_0}$.
\end{itemize}
It can be verified that the relation $\mathcal{H}_x = \mathcal{G}_x - \mu (x - \xs)$ holds in all these examples. Note that it is important to ensure that $\mathcal{G}_x$ is practical. For example, the shifted stochastic gradient estimator $\phii{x} = [\pfi{x} - \pfi{\xs}] - \mu(x - \xs)$ does not induce a practical $\mathcal{G}_x$.

We also apply the idea of changing ``perspective'' to proximal operator $\tprox_i^\alpha$ as given below.
\begin{lemmas}[Shifted firm non-expansiveness]\label{p-point-contract}
	Given relations $z^+ = \tprox^\alpha_{i} (z^-)$ and $y^+ = \tprox^\alpha_{i} (y^-)$, it holds that
	\[
	\frac{1}{\alpha^2}\left(1 + \frac{2(\alpha + \mu)}{L - \mu}\right) \norm{\phii{z^+} - \phii{y^+}}^2 + \left(1 + \frac{\mu}{\alpha}\right)^2 \norm{z^+ - y^+}^2 \leq  \norm{z^- - y^-}^2.
	\]
\end{lemmas}

\begin{lemma-remark}
	Recall the definition of a firmly non-expansive operator $T$ (e.g., Definition 4.1 in the textbook \citep{bauschke2017convex}): $\forall x, y$,
	$
	\norm{Tx - Ty}^2 + \norm{(\textup{Id} - T)x - (\textup{Id} - T)y}^2 \leq \norm{x - y}^2.
	$
	Lemma \ref{p-point-contract} can be derived by choosing\footnote{In the strongly convex setting, $(1 + \frac{\mu}{\alpha}) \cdot \tprox^\alpha_i$ is firmly non-expansive (e.g., Proposition 1 in \citet{point-SAGA}).} $T = (1 + \frac{\mu}{\alpha}) \cdot \tprox^\alpha_i$ and strengthening $\innr{Tx - Ty, (\textup{Id} - T)x - (\textup{Id} - T)y} \geq 0$ using the interpolation condition. A similar lemma has also been used in the analysis of the proximal point algorithm \citep{rockafellar1976monotone}. In our problem setting, \citet{point-SAGA} also strengthened firm non-expansiveness, which produces a $(1 + \frac{\mu}{\alpha})^{-1}$ contraction ratio instead of the above $(1 + \frac{\mu}{\alpha})^{-2}$ ratio created by shifting objective.
\end{lemma-remark}

Now we have all the building blocks to migrate existing schemes to tackle the shifted objective. To maximize the potential of our methodology, we focus on developing accelerated methods. We can also tighten the analysis of non-accelerated methods, which could lead to new algorithmic schemes.

\section{Deterministic Objectives}\label{sec:deterministic}

We consider the objective function \eqref{prob_def} with $n=1$. To begin, we recap the guarantee of NAG to facilitate the comparison. The proof is given in Appendix \ref{app-NAG} for completeness. At iteration $K-1$, NAG produces
\[
f(x_{K}) - f(\xs) + \frac{\mu}{2}\norm{z_{K} - \xs}^2 \leq \left(1 - \frac{1}{\sqrt{\kappa}}\right)^K\left(f(x_0) - f(\xs) + \frac{\mu}{2}\norm{z_0 - \xs}^2\right),
\]
where $x_0, z_0\in \R^d$ are the initial guesses. Denote the initial constant as $C^{\text{NAG}}_0 \triangleq f(x_0) - f(\xs) + \frac{\mu}{2}\norm{z_0 - \xs}^2$. This guarantee shows that in terms of reducing $\norm{x - \xs}^2$ to $\epsilon$, the sequences $\{x_k\}$ (due to $f(x_K) - f(\xs) \geq \frac{\mu}{2} \norm{x_K - \xs}^2$) and $\{z_k\}$
have the same iteration complexity $\sqrt{\kappa} \log{\frac{2C_0^{\text{NAG}}}{\mu\epsilon}}$. 


\subsection{Generalized Triple Momentum Method}
\label{sec:G-TM}
We present the first application of our methodology in Algorithm~\ref{G-TM}, which can be regarded as a technical migration\footnote{In our opinion, the most important techniques in NAG are Lemma \ref{func-contract} for $f$ and the mirror descent lemma. Algorithm \ref{G-TM} was derived by having a shifted version of Lemma \ref{func-contract} for $h$ and the shifted mirror descent lemma.} of NAG to the shifted objective. It turns out that Algorithm \ref{G-TM}, when tuned optimally, is equivalent to TM \citep{TM} (except for the first iteration). We thus name it as Generalized Triple Momentum method (G-TM). In comparison with TM, G-TM has the following advantages:
\begin{itemize}[leftmargin=*]

	
	\item \textit{Refined convergence guarantee.} TM has the guarantee
	(Eq.(11) in \citet{RMM} with $\rho = 1 - \frac{1}{\sqrt{\kappa}}$):
	\[
	\norm{z_K - \xs}^2 \leq \left(1 - \frac{1}{\sqrt{\kappa}}\right)^{2(K-1)} \left(\norm{z_1 - \xs}^2 + \frac{L-\mu}{L\mu}\left(h(y_{0}) - \frac{1}{2(L - \mu)} \norm{\ph{y_{0}}}^2\right)\right),
	\]
	which has an initial state issue: its initial constant correlates with $z_1$, which is not an initial guess. It can be verified that the first iteration of TM is GD with a $\frac{1}{\sqrt{L\mu}}$ stepsize, which exceeds the $\frac{2}{L + \mu}$ limit, and thus we do not have $\norm{z_1 - \xs}^2 \leq \norm{z_0 - \xs}^2$ in general. This issue is possibly the reason for the $\log{\sqrt{\kappa}}$ factor stated in \citet{TM}. G-TM resolves this issue and removes the log factor.
	
	\item \textit{More extensible proof.}  Our proof of G-TM is based on Lemma \ref{p-dis-contract}, which, as mentioned in Section~\ref{sec:lemmas}, allows shifted stochastic gradients. In comparison, the analysis of TM starts with establishing an algebraic identity and it is unknown whether this identity holds in the stochastic case.
	
	
	\item \textit{General scheme.} The framework of G-TM covers both NAG and TM (Appendix~\ref{app:unified}). When $\mu = 0$, it also covers the optimized gradient method \citep{OGM}, which is discussed in Section \ref{sec:conclusion}. 
\end{itemize}
\begin{algorithm}[t]
	\caption{Generalized Triple Momentum (G-TM)}
	\label{G-TM}
	\renewcommand{\algorithmicrequire}{\textbf{Input:}}
	\renewcommand{\algorithmicensure}{\textbf{Output:}}
	\begin{algorithmic}[1]
		\REQUIRE $\{\alpha_k>0\} , \{\tau^x_k \in ]0, 1[\}, \{\tau^{z}_k > 0\}$, initial guesses $y_{-1}, z_0 \in \R^d$ and iteration number $K$.
		\FOR{$k=0,\ldots, K-1$} 
		\STATE $y_k = \tau^x_k z_k + (1 - \tau^x_k) y_{k-1} + \tau^{z}_k \big(\mu(y_{k-1} - z_k)-\pf{y_{k-1}}\big)$.
		\STATE $z_{k+1} = \arg\min_{x} \Big\{\innr{\pf{y_k}, x} + (\alpha_k/2) \norm{x - z_k}^2 + (\mu/2)	\norm{x - y_k}^2 \Big\}$.
		\ENDFOR
		\ENSURE $z_K$.
	\end{algorithmic}
\end{algorithm}

A subtlety of Algorithm \ref{G-TM} is that it requires storing a past gradient vector, and thus at the first iteration, two gradient computations are needed. The analysis of G-TM is based on the same Lyapunov function in \citet{RMM}: \[T_k = h(y_{k-1}) - \frac{1}{2(L - \mu)} \norm{\ph{y_{k-1}}}^2 + \frac{\lambda}{2} \norm{z_k - \xs}^2\text{, where }\lambda >0.\]In the following theorem, we establish the per-iteration contraction of G-TM and the proof is given in Appendix \ref{app:proof_of_GTM}.


\begin{theorems}\label{GTM-contract}
	In Algorithm~\ref{G-TM}, if we fix $\tau^z_k = \frac{1 - \tau^x_k}{L - \mu}, \forall k$ and choose $\{\alpha_k\}, \{\tau^x_k\}$ under the constraints \[2\alpha_k \geq L\tau^x_k - \mu\text{ and }\left(1 + \frac{\mu}{\alpha_k}\right)^2 (1 - \tau^x_k) \leq 1, \]the iterations satisfy the contraction $T_{k+1} \leq (1 + \frac{\mu}{\alpha_k})^{-2} T_k$ with $\lambda = \frac{(\tau^x_k - \mu \tau^z_k)(\alpha_k + \mu)^2}{\alpha_k}$.
\end{theorems}

When the constraints hold as equality, we derive a simple constant choice for G-TM: $\alpha = \sqrt{L\mu} - \mu, \tau_x = \frac{2\sqrt{\kappa} - 1}{\kappa}, \tau_z = \frac{\sqrt{\kappa} - 1}{L(\sqrt{\kappa} + 1)}$. Here we also provide the parameter choices of NAG and TM under the framework of G-TM for comparison. Detailed derivation is given in Appendix \ref{app:unified}.
\[
\begin{aligned}
	&\text{NAG}\begin{cases}
		\alpha = \sqrt{L\mu} - \mu;\\
		\tau^x_k = (\sqrt{\kappa} + 1)^{-1}, \tau^z_k = 0, &k = 0; \\ \tau^x_k = (\sqrt{\kappa})^{-1}, \tau^z_k = \frac{1}{L+\sqrt{L\mu}}, &k \geq 1.
	\end{cases}	 &&&\text{TM}\begin{cases}
		\alpha = \sqrt{L\mu} - \mu;\\
		\tau^x_k = (\sqrt{\kappa} + 1)^{-1}, \tau^z_k = 0, &k = 0; \\ \tau^x_k = \frac{2\sqrt{\kappa} - 1}{\kappa}, \tau^z_k = \frac{\sqrt{\kappa} - 1}{L(\sqrt{\kappa} + 1)}, &k \geq 1.
	\end{cases}	
\end{aligned}
\]
Using the constant choice in Theorem \ref{GTM-contract}, by telescoping the contraction from iteration $K-1$ to $0$, we obtain
\begin{equation}\label{GTM-final}
	\frac{\mu}{2} \norm{z_K - \xs}^2 \leq \left(1 - \frac{1}{\sqrt{\kappa}}\right)^{2K} \left(\frac{\kappa - 1
	}{2\kappa}\left(h(y_{-1}) - \frac{1}{2(L - \mu)} \norm{\ph{y_{-1}}}^2\right) + \frac{\mu}{2} \norm{z_0 - \xs}^2\right).
\end{equation}
Denoting the initial constant as $C^{\text{G-TM}}_0 \triangleq \frac{\kappa - 1
}{2\kappa}(h(y_{-1}) - \frac{1}{2(L - \mu)} \norm{\ph{y_{-1}}}^2) + \frac{\mu}{2} \norm{z_0  - \xs}^2$, if we align the initial guesses $y_{-1} = x_0$ with NAG, we have $C^{\text{G-TM}}_0 \ll C^{\text{NAG}}_0$. This guarantee yields a $\frac{\sqrt{\kappa}}{2} \log{\frac{2C_0^{\text{G-TM}}}{\mu\epsilon}}$ iteration complexity for G-TM, which is at least two times lower than that of NAG and does not suffer from an additional $\log{\sqrt{\kappa}}$ factor as is the case for the original TM.

\subsubsection{The Tightness of \eqref{GTM-final}}
\label{sec:tightness}
It is natural to ask how tight the worst-case guarantee \eqref{GTM-final} is. We show that for the quadratic\footnote{This is also the example where GD with $\frac{2}{L + \mu}$ stepsize behaves exactly like its worst-case analysis.} $f(x) = \frac{1}{2} \innr{D^{\kappa} x,x}$ where $D^{\kappa}\triangleq\text{diag}(L,\mu)$ is a diagonal matrix, G-TM converges exactly at the rate in \eqref{GTM-final}. Note that for this objective, $h(x) - \frac{1}{2(L - \mu)} \norm{\ph{x}}^2 \equiv 0$, which means that the guarantee becomes
\[\norm{z_K - \xs}^2 \leq \left(1 - \frac{1}{\sqrt{\kappa}}\right)^{2K} \norm{z_0 - \xs}^2.\] Expanding the recursions in Algorithm \ref{G-TM}, we obtain the following result and its proof is given in Appendix~\ref{app:GTM-example}.
\begin{propositions}\label{GTM-example}
	If $f(x) =  \frac{1}{2} \innr{D^{\kappa} x,x}$, G-TM produces
	$
	\norm{z_{K} - \xs}^2 = \left(1 - \frac{1}{\sqrt{\kappa}}\right)^{2K} \norm{z_0 - \xs}^2.
	$
\end{propositions}

\section{Finite-Sum Objectives with Incremental First-Order Oracle}\label{sec:variance-reduction}

We now consider the finite-sum objective \eqref{prob_def} with $n \geq 1$. We choose SVRG \citep{SVRG} as the base algorithm to implement our boosting technique, and we also show that an accelerated SAGA \citep{SAGA} variant can be similarly constructed in Section \ref{sec:TM-SAGA}. Proofs in this section are given in Appendix~\ref{app:proofs_of_vr}.

\subsection{BS-SVRG}
\label{sec:TM-SVRG}
As mentioned in Section \ref{sec:lemmas}, the shifted SVRG estimator $\mathcal{H}^{\text{SVRG}}_x$ induces a practical $\mathcal{G}^{\text{SVRG}}_x$ (which is just the original SVRG estimator \citep{SVRG}) and thus by using Lemma \ref{p-dis-contract}, we obtain a practical updating rule and a classic equality for the shifted estimator. Now we can design an accelerated SVRG variant that minimizes $h$. To make the notations specific, we define
$
\mathcal{G}^{\text{SVRG}}_{x_k} \triangleq \pfik{x_k} - \pfik{\tilde{x}_s} + \pf{\tilde{x}_s},
$
where $i_k$ is sampled uniformly in $[n]$ and $\tilde{x}_s$ is a previously chosen random anchor point. 
For simplicity, in what follows, we only consider constant parameter choices. We name our SVRG variant as BS-SVRG (Algorithm \ref{TM-SVRG}), which is designed based on the following thought experiment.

\begin{algorithm}[t]
	\caption{SVRG Boosted by Shifting objective (BS-SVRG)}
	\label{TM-SVRG}
	\renewcommand{\algorithmicrequire}{\textbf{Input:}}
	\renewcommand{\algorithmicensure}{\textbf{Initialize:}}
	\begin{algorithmic}[1]
		\REQUIRE Parameters $\alpha > 0, \tau_x \in ]0, 1[$, initial guess $x_0 \in \R^d$, epoch number $S$ and epoch length $m$.
		\ENSURE Vectors $z^0_0 = \tilde{x}_0 = x_0$, constants $\tau_z = \frac{\tau_x}{\mu} - \frac{\alpha(1 - \tau_x)}{\mu(L-\mu)}, \widetilde{\omega} = \sum_{k=0}^{m-1}{\left(1 + \frac{\mu}{\alpha}\right)^{2k}}$.
		\FOR{$s=0, \ldots, S-1$}
		\STATE Compute and store $\pf{\tilde{x}_s}$.
		\FOR{$k=0,\ldots, m-1$} 
		\STATE $y^s_{k} = \tau_x z^s_k + \left(1 - \tau_x\right) \tilde{x}_s +  \tau_{z}\left(\mu(\tilde{x}_s - z^s_k) - \pf{\tilde{x}_s}\right)$.
		\STATE $z^s_{k+1} = \arg\min_{x} \left\{\innr{\mathcal{G}^{\text{SVRG}}_{y_k^s}, x} + (\alpha / 2) \norm{x - z^s_k}^2 + (\mu/2) \norm{x - y^s_k}^2 \right\}$.
		\ENDFOR
		\STATE $\tilde{x}_{s+1}$ is sampled from $\left\{ P(\tilde{x}_{s+1} = y_k^s) = \frac{1}{\widetilde{\omega}}\left(1 + \frac{\mu}{\alpha}\right)^{2k} \cond k\in \{0, \ldots, m-1\} \right\}.$
		\STATE $z^{s+1}_0 = z^s_{m}.$
		\ENDFOR
		\renewcommand{\algorithmicensure}{\textbf{Output:}}
		\ENSURE $z^{S}_0$.
	\end{algorithmic}
\end{algorithm}

{\bf\noindent Thought experiment.} We design BS-SVRG by extending G-TM, which is natural since almost all the existing stochastic accelerated methods are constructed based on NAG. For SVRG, its (directly) accelerated variants \citep{zhu:Katyusha,kw:MiG,lan2019unified} all incorporate the idea of ``negative'' momentum, which is basically Nesterov's momentum provided by the anchor point $\tilde{x}_s$ instead of the previous iterate. Inspired by their success, we design the ``momentum step'' of BS-SVRG (Step 4) by replacing all the previous iterate $y_{k-1}$ in $y_k = \tau_x z_k + (1 - \tau_x) y_{k-1} + \tau_z \big(\mu(y_{k-1} - z_k)-\pf{y_{k-1}}\big)$ with the anchor point $\tilde{x}_s$. The insight is that the ``momentum step'' is aggressive and could be erroneous in the stochastic case. Thus, we construct it based on some ``stable'' point instead of the previous stochastic iterate. 


We adopt a similar Lyapunov function as G-TM: \[
T_s \triangleq  h(\tilde{x}_s) - c_1 \norm{\ph{\tilde{x}_s}}^2 + \frac{\lambda}{2} \norm{z^s_0 - \xs}^2
\text{, where }c_1 \in \left[0, \frac{1}{2(L-\mu)}\right]\text{ and }\lambda > 0,\] and  
build the per-epoch contraction of BS-SVRG as follows.
\begin{theorems}\label{TM-SVRG-contract}
	In Algorithm~\ref{TM-SVRG}, if we choose $\alpha, \tau_x$ under the constraints \[\left(1 + \frac{\mu}{\alpha}\right)^{2m} (1 - \tau_x) \leq 1\text{ and }(1 + \tau_x)^2(1 - \tau_x)\geq 4\left(\left(\frac{\alpha}{\mu} + 1\right) - \left(\frac{\alpha}{\mu} + \kappa\right)\tau_x\right)^2, \]the per-epoch contraction $\E{T_{s+1}} \leq (1 + \frac{\mu}{\alpha})^{-2m} T_s$ holds with $\lambda = \frac{\alpha^2 (1 - \tau_x)}{\widetilde{\omega}(L - \mu)}(1 + \frac{\mu}{\alpha})^{2m}$. The expectation is taken with respect to the information up to epoch $s$.
\end{theorems}

In what follows, we provide a simple analytic choice that satisfies the constraints. We consider the ill-conditioned case where $\frac{m}{\kappa} \leq \frac{3}{4}$, and we fix $m=2n$ to make it specific.\footnote{We choose the setting that is used in the analysis and experiments of Katyusha \citep{zhu:Katyusha} to make a fair comparison.} In this case, \citet{zhu:Katyusha} derived an $O(\sqrt{6n\kappa} \log{\frac{1}{\epsilon}})$ expected iteration complexity\footnote{We are referring to the expected number of stochastic iterations (e.g., in total $Sm$ in Algorithm \ref{TM-SVRG}) required to achieve $\norm{x - \xs}^2 \leq \epsilon$. If $m=2n$, in average, each stochastic iteration of SVRG requires $1.5$ oracle calls.} for Katyusha (cf. Theorem 2.1, \citep{zhu:Katyusha}).

\begin{propositions}[Ill condition]\label{TM-SVRG-choice}
	If $\frac{m}{\kappa} \leq \frac{3}{4}$, the choice $\alpha = \sqrt{cm\mu L} - \mu,
	\tau_x = (1 - \frac{1}{c\kappa})\frac{\sqrt{cm\kappa}}{\sqrt{cm\kappa} + \kappa - 1}$, where $c = 2 + \sqrt{3}$, satisfies the constraints in Theorem \ref{TM-SVRG-contract}.
\end{propositions}

Using this parameter choice in Theorem \ref{TM-SVRG-contract}, we obtain an $O(\sqrt{1.87 n\kappa }\log{\frac{1}{\epsilon}})$ expected iteration complexity for BS-SVRG, which is around $1.8$ times lower than that of Katyusha.  
\begin{remark}
	We are not aware of other parameter choices of Katyusha that have faster rates. \citet{hu18b} made an attempt based on dissipativity theory, but no explicit rate is given. To derive a better choice for Katyusha, significant modification to its proof is required (for its parameter $\tau_2$), which results in complicated constraints and is thus out of the scope of this paper. We believe that there could be some computer-aided ways to find better choices for both Katyusha and BS-SVRG, which we leave for future work.
\end{remark}

For the other case where $\frac{m}{\kappa} > \frac{3}{4}$ (i.e., $\kappa = O(n)$), almost all the accelerated and non-accelerated incremental gradient methods perform the same, at an $O(n\log{\frac{1}{\epsilon}})$ oracle complexity (and is indeed fast). \citet{hannah2018breaking} shows that by optimizing the parameters of SVRG and SARAH, a lower $O(n + \frac{n}{1+\max{\{\log{(n/\kappa)}, 0\}}} \log{\frac{1}{\epsilon}})$ oracle complexity is achievable. Due to these facts, we do not optimize the parameters for this case and provide the following proposition as a basic guarantee.

\begin{propositions}[Well condition]\label{TM-SVRG-choice-well}
	If $\frac{m}{\kappa} > \frac{3}{4}$, by choosing $\alpha = \frac{3L}{2} - \mu, \tau_x = (1 - \frac{1}{6m}) \frac{3\kappa}{5\kappa - 2}$, the epochs of BS-SVRG satisfy $T_{s+1} \leq \frac{1}{2}\cdot T_{s}$ with $\lambda = \frac{2\alpha^2 (1 - \tau_x)}{\widetilde{\omega}(L - \mu)}$, which implies an $O(n\log{\frac{1}{\epsilon}})$ expected iteration complexity.
\end{propositions}

There exists a special choice in the constraints: by choosing $\tau_x = \frac{\alpha + \mu}{\alpha + L}$, the second constraint always holds and this leads to $c_1 = 0$ in $T_s$. In this case, $\alpha$ can be found using numerical tools, which is summarized as follows.

\begin{propositions}[Numerical choice]\label{TM-SVRG-choice-numerical}
	By fixing $\tau_x = \frac{\alpha + \mu}{\alpha + L}$, the optimal choice of $\alpha$ can be found by solving the equation $\left(1 + \frac{\mu}{\alpha}\right)^{2m} \big(1 - \frac{\alpha + \mu}{\alpha + L}\big) = 1$ using numerical tools, and this equation has a unique positive root.
\end{propositions}

Compared with Katyusha, BS-SVRG has a simpler scheme, which only requires storing one variable vector $\{z_k\}$ and tuning $2$ parameters similar to MiG \citep{kw:MiG}. Moreover, BS-SVRG achieves the fastest rate among the accelerated SVRG variants.

\subsection{Accelerated SAGA Variant}
\label{sec:TM-SAGA}
As given in Section \ref{sec:lemmas}, the shifted SAGA estimator $\mathcal{H}^{\text{SAGA}}_x$ also induces a practical gradient estimator, and thus we can design an accelerated SAGA variant in a similar way.
Inspired by the existing (directly) accelerated SAGA variant \citep{kw:SSNM}, we can design the recursion (updating rule of the table) as $\phi_{i_k}^{k+1} = \tau_x z_k + \left(1 - \tau_x\right) \phi_{i_k}^{k} +  \tau_{z}\big(\mu (\frac{1}{n} \sum_{i = 1}^n{\phi^k_i} - z_k) - \frac{1}{n} \sum_{i=1}^n {\pfi{\phi^k_i}} \big)$. We found that for the resulting scheme, we can adopt the following Lyapunov function: \[T_k = \frac{1}{n}\sum_{i=1}^n{h_{i}(\phi^{k}_{i})} - c_1\norm{\frac{1}{n}\sum_{i=1}^n{\phii{\phi^k_{i}}}}^2 + \frac{\lambda}{2} \norm{z_k - \xs}^2\text{, where }c_1 \in \left[0, \frac{1}{2(L-\mu)}\right], \lambda > 0,\]which is an ``incremental version'' of $T_s$. Note that
\[
\frac{1}{n}\sum_{i=1}^n{h_{i}(\phi^{k}_{i})} - c_1\norm{\frac{1}{n}\sum_{i=1}^n{\phii{\phi^k_{i}}}}^2 \geq \frac{1}{n}\sum_{i=1}^n{\big(h_{i}(\phi^{k}_{i}) - c_1\norm{\phii{\phi^k_{i}}}^2\big)} \geq 0.
\]

A similar accelerated rate can be derived for the SAGA variant and its parameter choice shows some interesting correspondence between the variants of SVRG and SAGA. Moreover, the resulting scheme does not need the tricky ``doubling sampling'' in \citet{kw:SSNM} and thus it has a lower iteration complexity. However, since its updating rules require the knowledge of point table, the scheme has an undesirable $O(nd)$ memory complexity. We provide this variant in Appendix \ref{app:BS-SAGA} for interested readers.

\section{Finite-Sum Objectives with Incremental Proximal Point Oracle}\label{sec:proximal-point}

We consider the finite-sum objective \eqref{prob_def} and assume that the proximal operator oracle $\tprox_i^\alpha(\cdot)$ of each $f_i$ is available. Point-SAGA \citep{point-SAGA} is a typical method that utilizes this oracle, and it achieves the same $O\big((n + \sqrt{n\kappa})\log{\frac{1}{\epsilon}}\big)$ expected iteration complexity. Although in general, the incremental proximal operator oracle is much more expensive than the incremental gradient oracle, Point-SAGA is interesting in the following aspects: (1) it has a simple scheme with only $1$ parameter; (2) its analysis is elegant and tight, which does not require any Young's inequality; (3) for problems where the proximal point oracle has an analytic solution, it has a very fast rate (i.e., its rate factor is smaller than $1 - (n+\sqrt{n\kappa} + 1)^{-1}$, which is faster than both Katyusha and BS-SVRG). 

\begin{algorithm}[t]
	\caption{Point-SAGA Boosted by Shifting objective (BS-Point-SAGA)}
	\label{TM-Point-SAGA}
	\renewcommand{\algorithmicrequire}{\textbf{Input:}}
	\renewcommand{\algorithmicensure}{\textbf{Initialize:}}
	\begin{algorithmic}[1]
		\REQUIRE Parameters $\alpha > 0$ and initial guess $x_0 \in \R^d$, iteration number $K$.
		\ENSURE A point table $\phi^0 \in \R^{d\times n}$ with $\forall i\in [n], \phi^0_i = x_0$, running averages for the point table and its gradients.
		\FOR{$k=0, \ldots, K-1$}
		\STATE Sample $i_k$ uniformly in $[n]$.
		\STATE Update $x$: $z_k = x_k + \frac{1}{\alpha}\left(\pfik{\phi^k_{i_k}} - \frac{1}{n}\sum_{i=1}^n{\pfi{\phi^k_i}} + \mu \left(\frac{1}{n} \sum_{i = 1}^n{\phi^k_i} - \phi^k_{i_k}\right)\right)$, \\
		\qquad\quad\ \ \,\,\,\,\, $x_{k+1} = \tprox^\alpha_{i_k} (z_k).$
		\STATE Set $\phi_{i_k}^{k+1} = x_{k+1}$ and keep other entries unchanged (i.e., for $i\neq i_k, \phi^{k+1}_i = \phi^k_i$). Update the running averages according to the change in $\phi^{k+1}$ (note that $\pfik{\phi^{k+1}_{i_k}} = \alpha (z_k - x_{k+1})$).
		\ENDFOR
		\renewcommand{\algorithmicensure}{\textbf{Output:}}
		\ENSURE $x_K$.
	\end{algorithmic}
\end{algorithm}

It might be surprising that by shifting objective, the convergence rate of Point-SAGA can be further boosted. We name the proposed variant as BS-Point-SAGA, which is presented in Algorithm~\ref{TM-Point-SAGA}. Recall that the Lyapunov function used to analyze Point-SAGA has the form (cf. Theorem 5, \citep{point-SAGA}):
\[
T^{\text{Point-SAGA}}_k =  \frac{c}{n}\sum_{i=1}^n{\norm{\pfi{\phi^k_i} - \pfi{\xs}}^2} + \norm{x_k - \xs}^2.
\]
We adopt a shifted version of this Lyapunov function:
\[
T_k = \lambda \cdot \frac{1}{n}\sum_{i=1}^n{\norm{\phii{\phi^k_{i}}}^2} + \norm{x_k - \xs}^2\text{, where }\lambda >0.\] The analysis of BS-Point-SAGA is a direct application of Lemma \ref{p-point-contract}. We build the per-iteration contraction in the following theorem, and its proof is given in Appendix \ref{app:proof_of_point}.

\begin{theorems}\label{TM-Point-SAGA-contract} 
	In Algorithm \ref{TM-Point-SAGA}, if we choose $\alpha$ as the unique positive root of the cubic equation
	\[
	2\left(\frac{\alpha}{\mu}\right)^3 - \left(4n-6\right)\left(\frac{\alpha}{\mu}\right)^2 - \left(2n\kappa + 4n - 6\right) \left(\frac{\alpha}{\mu}\right) - \left(n\kappa + n - 2\right) = 0,
	\]
	the per-iteration contraction $\Eik{T_{k+1}} \leq (1 + \frac{\mu}{\alpha})^{-2} T_k$ holds with $\lambda = \frac{n}{\alpha^2} + \frac{2(\alpha + \mu)(n-1)}{\alpha^2(L - \mu)}$. The root of this cubic equation satisfies $\frac{\alpha}{\mu} = O(n + \sqrt{n\kappa})$, which implies an $O\big((n+\sqrt{n\kappa})\log{\frac{1}{\epsilon}}\big)$ expected iteration complexity.
\end{theorems}

\begin{wrapfigure}{r}{0.33\linewidth}
	\includegraphics[width=\linewidth]{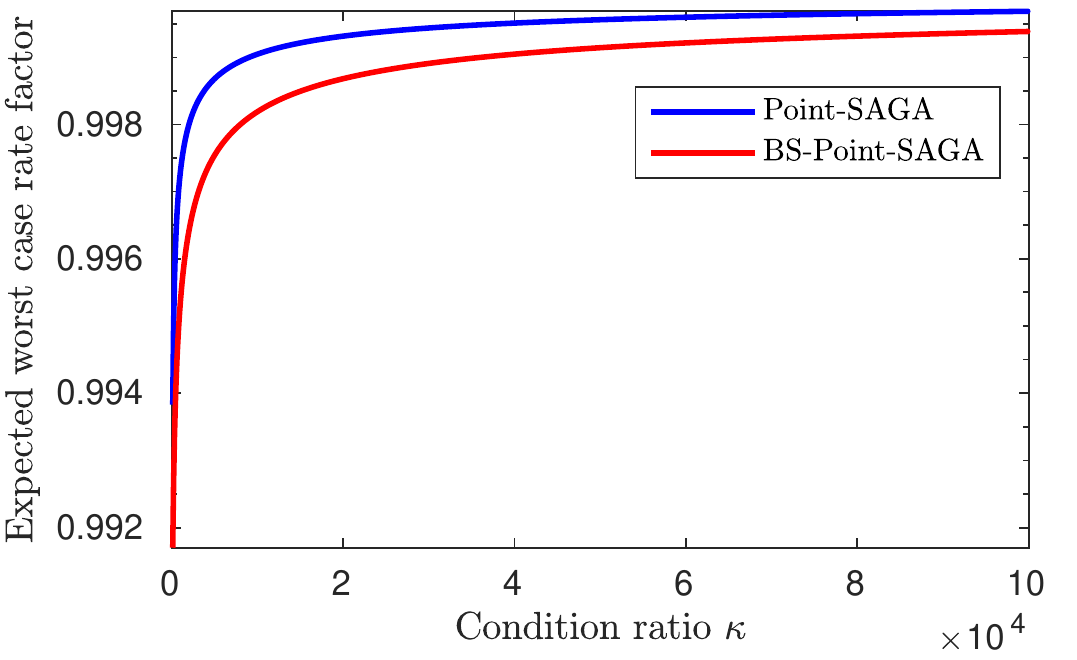}
	\caption{A comparison of the expected worst-case rate factors.}
	\label{point_worst_case_rate_factor}
\end{wrapfigure}

The expected worst-case rate factor of BS-Point-SAGA is minimized by solving the cubic equation in Theorem \ref{TM-Point-SAGA-contract} exactly. The analytic solution of this equation is messy, but it can be easily calculated using numerical tools. In Figure \ref{point_worst_case_rate_factor}, we numerically compare the rate factors of Point-SAGA and BS-Point-SAGA. When $\kappa$ is large, the rate factor of BS-Point-SAGA is close to the square of the rate factor of Point-SAGA, which implies an almost $2$ times lower expected iteration complexity. In terms of memory requirement, BS-Point-SAGA has an undesirable $O(nd)$ complexity since the update of $x_{k+1}$ involves $\phi^k_{i_k}$. Nevertheless, it achieves the fastest known rate for finite-sum problems (if both $L$ and $\mu$ are known), and we present it as a special instance of our design methodology.
\section{Performance Evaluations}\label{sec:evaluations}

In general, a faster worst-case rate does not necessarily imply a better empirical performance. It is possible that the slower rate is loose or the worst-case analysis is not representative of reality (e.g., worst-case scenarios are not stable to perturbations). We provide experimental results of the proposed methods in this section. We evaluate them in the ill-conditioned case where the problem has a huge $\kappa$ to justify the accelerated $\sqrt{\kappa}$ dependence. Detailed experimental setup can be found in Appendix \ref{app:exp}.
\begin{figure}[t]
	\begin{center}
		\begin{subfigure}{0.3174\linewidth}
			\includegraphics[width=\linewidth]{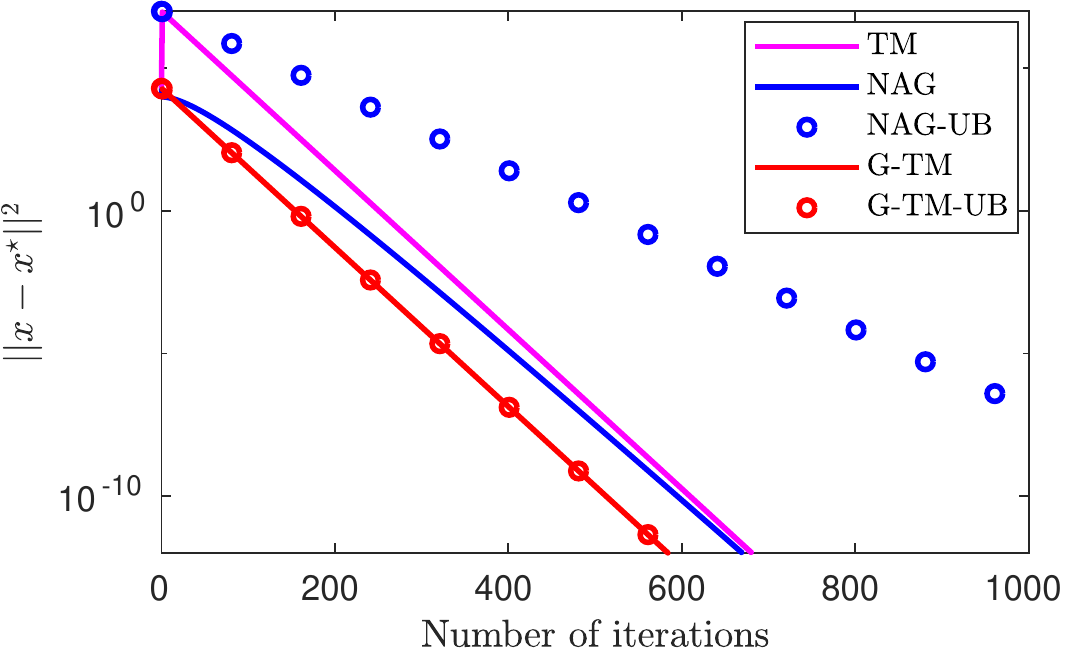}
			\caption{Simulation.}
			\label{Simu_deterministic}
		\end{subfigure}
		\begin{subfigure}{0.3174\linewidth}
			\includegraphics[width=\linewidth]{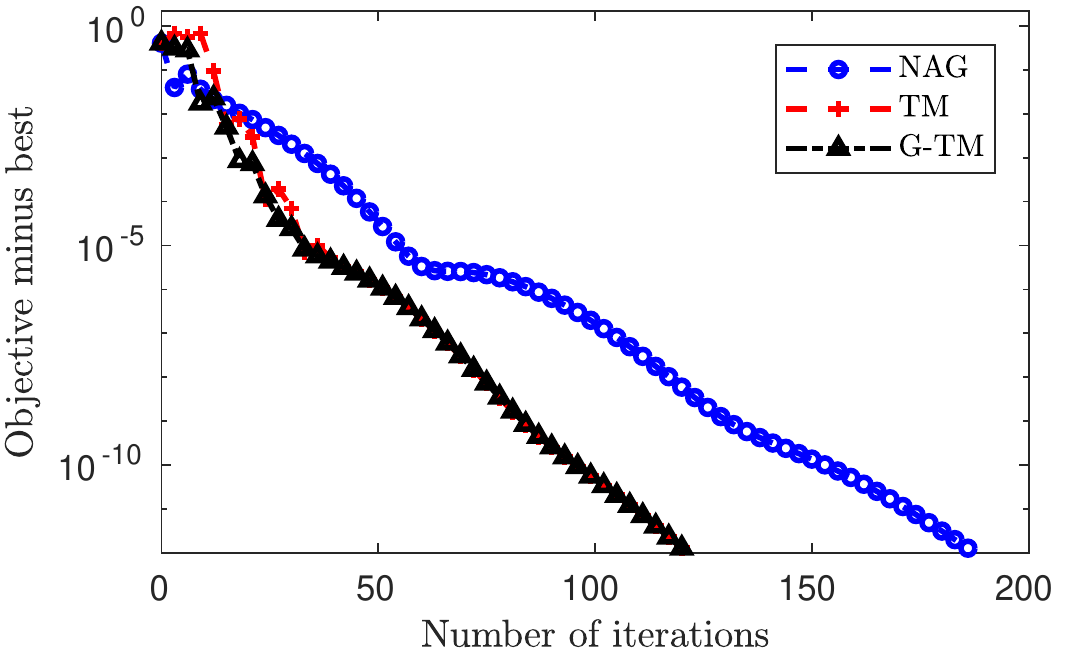}
			\caption{\textsf{ijcnn1} dataset.}
			\label{Eval_G_TM}
		\end{subfigure}
		\begin{subfigure}{0.3174\linewidth}
			\includegraphics[width=\linewidth]{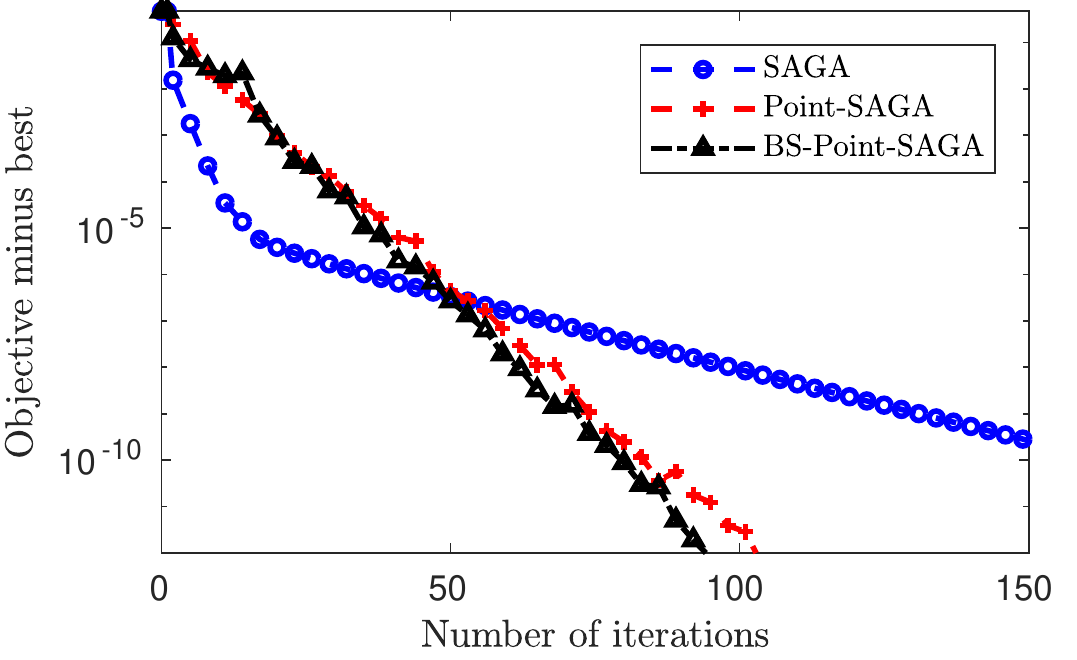}
			\caption{\textsf{w8a} dataset.}
			\label{Eval_point_SAGA}
		\end{subfigure}\\
		\begin{subfigure}{0.64\linewidth}
			\includegraphics[width=0.496\linewidth]{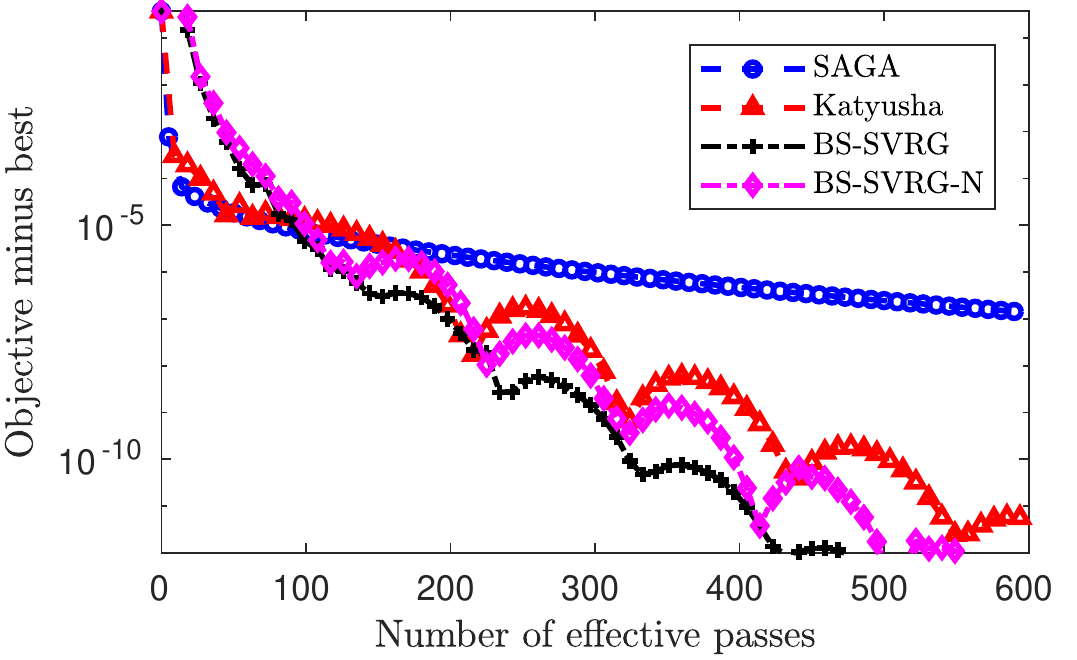}
			\includegraphics[width=0.496\linewidth]{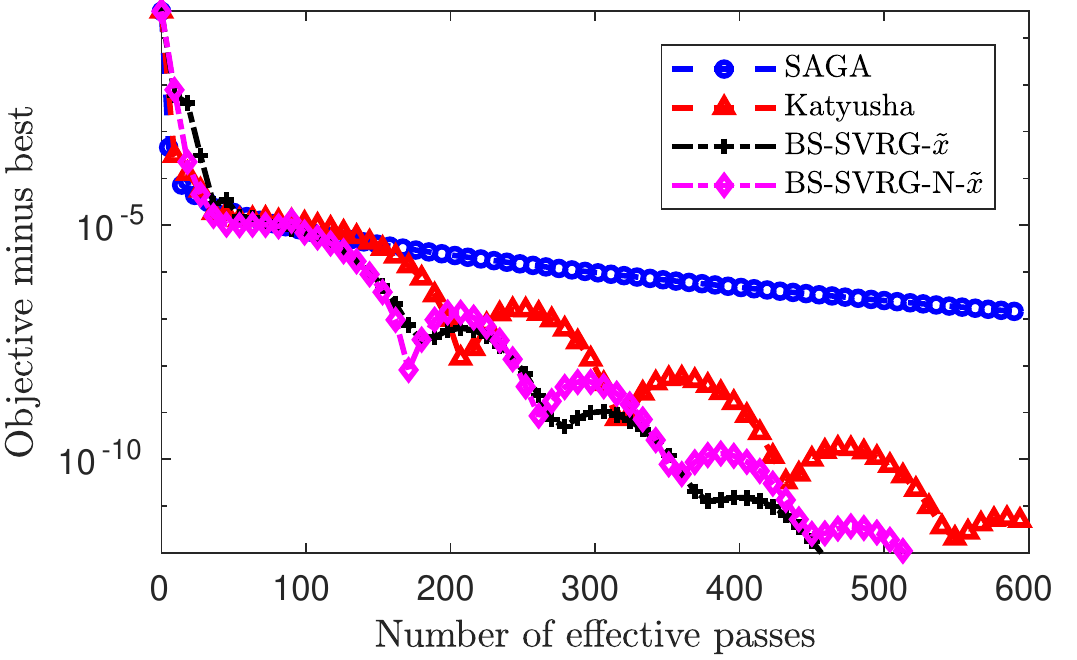}
			\caption{\textsf{a9a} dataset. BS-SVRG outputs $z$ (Left), outputs $\tilde{x}$ (Right).}
			\label{Eval_SVRG_a}
		\end{subfigure}
		\begin{subfigure}{0.3174\linewidth}
			\includegraphics[width=\linewidth]{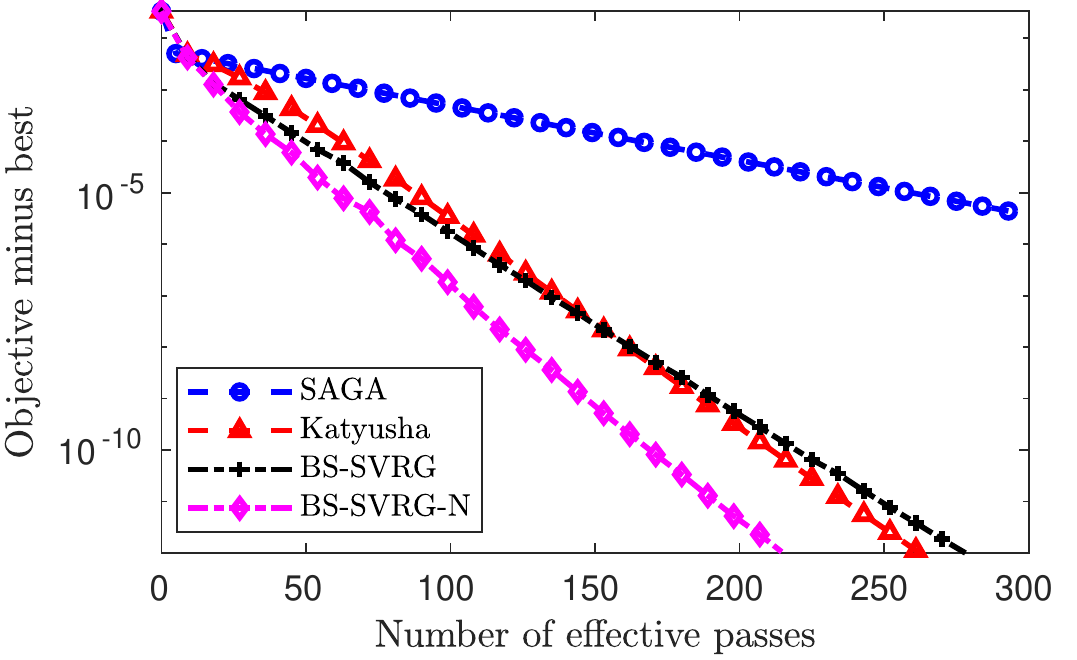}
			\caption{\textsf{covtype} dataset.}
			\label{Eval_SVRG_b}
		\end{subfigure}
		\caption{Evaluations. (a) Quadratic, $L=1, \mu = 10^{-3}$. (b) $\ell_2$-logistic regression, $\mu = 10^{-3}$. (c) Ridge regression, $\mu=5\times 10^{-7}$.
			(d) (e) $\ell_2$-logistic regression, $\mu=10^{-8}$.}
		\label{Eval_SVRG}
	\end{center}
\end{figure}

We started with evaluating the deterministic methods: NAG, TM and G-TM. We first did a simulation on the quadratic objective mentioned in Section \ref{sec:tightness}, which also serves as a justification of Proposition \ref{GTM-example}. In this simulation, the default (constant) parameter choices were used and all the methods were initialized in $(-100, 100)$. We plot their convergences and theoretical guarantees (marked with ``UB'') in Figure~\ref{Simu_deterministic} (the bound for TM is not shown due to the initial state issue). This simulation shows that after the first iteration, TM and G-TM have the same rate, and the initial state issue of TM can make it slower than NAG. It also  suggests that the guarantee of NAG is loose.

Then, we measured their performance on real world datasets from LIBSVM \citep{LIBSVM}. The task we chose is  $\ell_2$-logistic regression.
We normalized the datasets and thus for this problem, $L = 0.25 + \mu$. For real world tasks, we tracked function value suboptimality, which is easier to compute than $\norm{x - \xs}^2$ in practice. The result is given in Figure \ref{Eval_G_TM}. In the first $30$ iterations, TM is slower than G-TM due to the initial state issue. After that, they are almost identical and are faster than NAG.

We then evaluated BS-SVRG on the same problem, which can fully utilize the finite-sum structure. We evaluated two parameter choices of BS-SVRG: (1) the analytic choice in Proposition \ref{TM-SVRG-choice} (marked as \mbox{``BS-SVRG''}); (2) the numerical choice in Proposition \ref{TM-SVRG-choice-numerical} (marked as``BS-SVRG-N''). We selected SAGA ($\gamma = \frac{1}{2(\mu n + L)}$, \citep{SAGA}) and Katyusha ($\tau_2 = \frac{1}{2}, \tau_1 = \sqrt{\frac{m}{3\kappa}}, \alpha = \frac{1}{3\tau_1 L}$, \citep{zhu:Katyusha}) with their default parameter choices as the baselines. Since SAGA and SVRG-like algorithms have different iteration complexities, we plot the curve with respect to the number of data passes. The results are given in Figure \ref{Eval_SVRG_a} and \ref{Eval_SVRG_b}. In the experiment on \textsf{a9a} dataset (Figure~\ref{Eval_SVRG_a}~(Left)), both choices of BS-SVRG perform well after $100$ passes. The issue of their early stage performance can be eased by outputting the anchor point $\tilde{x}$ instead, as shown in Figure~\ref{Eval_SVRG_a}~(Right).

We also conducted an empirical comparison between BS-Point-SAGA and Point-SAGA in Figure~\ref{Eval_point_SAGA}. Their analytic parameter choices were used. We chose ridge regression as the task since its proximal operator has a closed form solution (see Appendix A in \citet{point-SAGA}). For this objective, after normalizing the dataset, $L = 1 + \mu$. The performance of SAGA is also plotted as a reference.

\section{Conclusion}\label{sec:conclusion}
In this work, we focused on unconstrained smooth strongly convex problems and designed new schemes for a shifted objective. Lemma \ref{p-dis-contract} and Lemma \ref{p-point-contract} are the cornerstones for the new designs, which serve as instantiations of the shifted gradient oracle. Following this methodology, we proposed G-TM, BS-SVRG (and BS-SAGA) and BS-Point-SAGA. The new schemes achieve faster worst-case rates and have tighter and simpler proofs compared with their existing counterparts. Experiments on machine learning tasks show some improvement of the proposed methods. 

Although provided only for strongly convex problems, our framework of exploiting the interpolation condition (i.e., Algorithm \ref{G-TM}) can also be extended to the non-strongly convex case ($\mu=0$). It can be easily verified that Theorem \ref{GTM-contract} holds with $\mu=0$ and thus we can choose a variable-parameter setting that leads to the $O(1/K^2)$ rate. It turns out that Algorithm \ref{G-TM} in this case is equivalent to the optimized gradient method \citep{OGM}, which is also covered by the second accelerated method (14) studied in \citet{computer-aided}. Moreover, the Lyapunov function $T_k$ becomes $a_k\big(f(y_{k-1}) - f(\xs) - \frac{1}{2L}\norm{\pf{y_{k-1}}}^2\big) + \frac{L}{4}\norm{z_k - \xs}^2$ for some $a_k> 0$, which is exactly the one used in Theorem 11, \citep{computer-aided}.

While the proposed approach boosts the convergence rate, some limitations should be stressed. First, it requires a prior knowledge of the strong convexity constant $\mu$ since even if it is applied to a non-accelerated method, the parameter choice is always related to $\mu$. Furthermore, this methodology relies heavily on the interpolation condition, which requires $f$ to be defined everywhere on $\R^d$ \citep{drori2018properties}. This restriction makes it hardly generalizable to the constrained/proximal setting \citep{nesterov2013gradient} (for the proximal case, a possible solution is to assume that the smooth part is defined everywhere on $\R^d$ \citep{FISTA,kim2018another,taylor2017exact}).

\bibliographystyle{apalike}
\bibliography{TM}

\newpage
\appendix

\section{Technical lemmas with proofs}
\label{app:lemmas}

\begin{app-lemmas}[Shifted mirror descent lemma]\label{dis-contract}
	Given a gradient estimator $\mathcal{G}_y$, vectors $z^+, z^-, y\in \R^d$, fix the updating rule $z^+ = \arg\min_{x} \big\{\innr{\mathcal{G}_y, x} + \frac{\alpha}{2} \norm{x - z^-}^2 + \frac{\mu}{2}	\norm{x - y}^2 \big\}$. Suppose that we have a shifted gradient estimator $\mathcal{H}_y$ satisfying the relation $\mathcal{H}_y = \mathcal{G}_y - \mu (y - \xs)$, it holds that
	\[
	\innr{\mathcal{H}_y, z^- - \xs} = \frac{\alpha}{2} \left(\norm{z^- - \xs}^2 - \left(1 + \frac{\mu}{\alpha}\right)^2\norm{z^+ - \xs}^2\right) + \frac{1}{2\alpha} \norm{\mathcal{H}_y}^2.
	\]
	\begin{proof}
		Using the optimality condition,
		\[
		\begin{gathered}
			\mathcal{G}_y + \alpha(z^+ - z^-) + \mu (z^+ - y) = \mathbf{0}, \\
			\mathcal{H}_y + \alpha(z^+ - z^-) + \mu (z^+ - \xs) = \mathbf{0}, \\
			(\alpha + \mu) (z^+ - \xs) = \alpha (z^- - \xs) - \mathcal{H}_y, \\
			(\alpha + \mu)^2 \norm{z^+ - \xs}^2 = \alpha^2 \norm{z^- - \xs}^2 - 2\alpha \innr{\mathcal{H}_y, z^- - \xs} + \norm{\mathcal{H}_y}^2.
		\end{gathered}
		\]
		Re-arranging the last equality completes the proof.
	\end{proof}
\end{app-lemmas}

\begin{app-lemmas}[Shifted firm non-expansiveness]\label{point-contract}
	Given relations $z^+ = \tprox^\alpha_{i} (z^-)$ and $y^+ = \tprox^\alpha_{i} (y^-)$, it holds that
	\[
	\frac{1}{\alpha^2}\left(1 + \frac{2(\alpha + \mu)}{L - \mu}\right) \norm{\phii{z^+} - \phii{y^+}}^2 + \left(1 + \frac{\mu}{\alpha}\right)^2 \norm{z^+ - y^+}^2 \leq  \norm{z^- - y^-}^2.
	\]
	\begin{proof}
		Based on the first-order optimality condition and the definition of $h_i$,
		\[
		\begin{gathered}
			\pfi{z^+} + \alpha (z^+ - z^-) = \mathbf{0},\qquad \pfi{y^+} + \alpha (y^+ - y^-) = \mathbf{0}, \\
			\phii{z^+} + \pfi{\xs} + \mu(z^+ - \xs) + \alpha (z^+ - z^-) = \mathbf{0},\\
			\phii{y^+} + \pfi{\xs} + \mu(y^+ - \xs) + \alpha (y^+ - y^-) = \mathbf{0}.
		\end{gathered}
		\]
		
		Subtract the last two equalities,
		\begin{equation}\label{point-P1}
			(\alpha + \mu) (z^+ - y^+) = \alpha (z^- - y^-) - \big(\phii{z^+} - \phii{y^+}\big),
		\end{equation}
		which implies
		\begin{equation}\label{point-P2}
			\begin{aligned}
				(\alpha + \mu)^2 \norm{z^+ - y^+}^2 ={}& \alpha^2 \norm{z^- - y^-}^2 - 2\alpha\innr{\phii{z^+} - \phii{y^+}, z^- - y^-} \\&+ \norm{\phii{z^+} - \phii{y^+}}^2.
			\end{aligned}
		\end{equation}
		
		Based on the interpolation condition of $h_i$, we have
		\[
		\innr{\phii{z^+} - \phii{y^+}, z^+ - y^+} \geq \frac{1}{L - \mu} \norm{\phii{z^+} - \phii{y^+}}^2.
		\]
		
		Together with \eqref{point-P1}, it holds that
		\[
		\innr{\phii{z^+} - \phii{y^+}, z^- - y^-} \geq \frac{1}{\alpha}\left(1 + \frac{\alpha + \mu}{L - \mu}\right) \norm{\phii{z^+} - \phii{y^+}}^2.
		\]
		
		It remains to use this bound in \eqref{point-P2}. 
	\end{proof}
\end{app-lemmas}

Forming convex combination between vector sequences is a common technique in designing accelerated methods (e.g., \cite{AT-NAG, lan2012optimal, ghadimi2012optimal, zhu:Katyusha}). From an analytical perspective, convex combination facilitates building a contraction between function values and the coefficient directly controls the contraction ratio, which is summarized in the following lemma. Unlike previous works, we allow a residual term $\mathcal{R}$ in the convex combination.

\begin{app-lemmas}[Function-value contraction]\label{func-contract}
	Given a continuously differentiable and convex function $f$, vectors $x^+, x^-, z, \mathcal{R}\in \R^d$ and scalar $\tau\in ]0,1[$, if $x^+ = \tau z + (1 - \tau)x^- + \mathcal{R}$, it satisfies that
	\[
	f(x^+) - f(\xs) \leq{} (1 - \tau)\big(f(x^-) - f(\xs)\big) + \innr{\pf{x^+}, \mathcal{R}}+ \tau\innr{\pf{x^+}, z - \xs}.
	\]
	
	\begin{proof} Using convexity twice,
		\[
		\begin{aligned}
			f(x^+) - f(\xs) \leq{}& \innr{\pf{x^+}, x^+ - \xs} \\={}& \innr{\pf{x^+}, x^+ - z} + \innr{\pf{x^+}, z - \xs} \\
			={}& \frac{1 - \tau}{\tau}\innr{\pf{x^+}, x^- - x^+} + \frac{1}{\tau} \innr{\pf{x^+}, \mathcal{R}}+ \innr{\pf{x^+}, z - \xs} \\
			\leq{}& \frac{1 - \tau}{\tau}\big(f(x^-) - f(x^+)\big) + \frac{1}{\tau} \innr{\pf{x^+}, \mathcal{R}}+ \innr{\pf{x^+}, z - \xs}.
		\end{aligned}
		\]
		Re-arranging this inequality completes the proof.
	\end{proof}
\end{app-lemmas}

This simple trick (with $\mathcal{R} = \mathbf{0}$) appears frequently in the proofs of existing accelerated first-order methods. Note that the convexity arguments in this lemma can be strengthened by the interpolation condition or strong convexity if $f$ satisfies additional assumptions.

\section{Proofs for Section \ref{sec:deterministic}}

\subsection{Generality of the framework of Algorithm \ref{G-TM}}
\label{app:unified}

First, we show that TM is a parameterization of NAG (Algorithm \ref{Uni_Acc} in Appendix \ref{app-NAG}). Note that TM has the following scheme (the notations follow the ones in \citet{RMM}):
\[
\begin{aligned}
	x_{k+1} &= x_k + \beta (x_k - x_{k-1}) - \alpha \pf{y_k}, \\
	y_{k+1} &= x_{k+1} + \gamma (x_{k+1} - x_k), \\
	z_{k+1} &= x_{k+1} + \delta (x_{k+1} - x_k).
\end{aligned}
\]

By casting this scheme into the framework of Algorithm \ref{Uni_Acc}, we obtain
\[
\begin{aligned}
	y_k &= \frac{\gamma}{\delta} z_k + \left(1 - \frac{\gamma}{\delta}\right) x_k, \\
	z_{k+1} &= \frac{\beta(1 + \delta) - \gamma}{\delta - \gamma} z_k +  \frac{\delta - \beta(1 + \delta)}{\delta - \gamma} y_k  - \alpha(1+\delta) \pf{y_k},\\
	x_{k+1} &= \frac{1}{1 + \delta} z_{k+1} + \frac{\delta}{1 + \delta} x_k.
\end{aligned}
\]

Substituting the parameter choice of TM, we see that TM is equivalent to choosing $\alpha = \sqrt{L\mu} - \mu, \tau_y = (\sqrt{\kappa} + 1)^{-1}, \tau_x = \frac{2\sqrt{\kappa} - 1}{\kappa}$ in Algorithm \ref{Uni_Acc}. Interestingly, this choice and the choice of NAG (given in Appendix~\ref{app-NAG}) only differ in $\tau_x$. 

Then, we show that Algorithm \ref{Uni_Acc} is an instance of the framework of Algorithm \ref{G-TM}. By expanding the convex combinations of sequences $\{y_k\}$ and $\{x_k\}$ in Algorithm \ref{Uni_Acc}, we can conclude that
\[
y_k = \tau_xz_k + (1 - \tau_x) y_{k-1} + \tau_y(1 - \tau_x) (z_k - z_{k-1}).
\]

Based on the optimality condition at iteration $k-1$, we have
\[
\alpha (z_k - z_{k-1}) = \mu(y_{k-1} - z_k) - \pf{y_{k-1}}.
\]

Now, it is clear that Algorithm \ref{Uni_Acc} is an instance of the framework of Algorithm \ref{G-TM} with the variable-parameter choice (let $y_{-1} = x_0$): at $k = 0, \tau^x_0 = \tau_y, \tau^z_0 = 0$; at $k \geq 1, \tau^x_k = \tau_x, \tau^z_k = \frac{\tau_y(1 - \tau_x) }{\alpha}$.

\subsection{Proof of Theorem \ref{GTM-contract}}
\label{app:proof_of_GTM}
First, we can introduce a contraction between $h(y_k)$ and $h(y_{k-1})$  using Lemma~\ref{func-contract}. Applying Lemma~\ref{func-contract} with $f=h$ for the recursion $y_k = \tau^x_k z_k + (1 - \tau^x_k) y_{k-1} + \tau^{z}_k \big(\mu(y_{k-1} - z_k)-\pf{y_{k-1}}\big)$ and strengthening the convexity arguments by the interpolation condition, we obtain 
\[
\begin{aligned}
	h(y_k) \leq{}& (1 - \tau^x_k)h(y_{k-1}) + \tau^z_k\innr{\ph{y_k},\mu(y_{k-1} - z_k)-\pf{y_{k-1}}}+ \tau^x_k\innr{\ph{y_k}, z_k - \xs} \\
	&- \frac{\tau^x_k}{2(L-\mu)} \norm{\ph{y_k}}^2 - \frac{1 - \tau^x_k}{2(L - \mu)} \norm{\ph{y_{k-1}} - \ph{y_k}}^2.
\end{aligned}
\]

Note that $\mu(y_{k-1} - z_k)-\pf{y_{k-1}} = \mu(\xs - z_k) - \ph{y_{k-1}}$ by definition, and thus
\begin{equation}\label{TM-h-contract}
	\begin{aligned}
		h(y_k) \leq{}& (1 - \tau^x_k)h(y_{k-1}) - \tau^z_k\innr{\ph{y_k},\ph{y_{k-1}}}+ (\tau^x_k - \mu\tau^z_k)\innr{\ph{y_k}, z_k - \xs} \\
		&- \frac{\tau^x_k}{2(L-\mu)} \norm{\ph{y_k}}^2 - \frac{1 - \tau^x_k}{2(L - \mu)} \norm{\ph{y_{k-1}} - \ph{y_k}}^2.
	\end{aligned}
\end{equation}

Then, to build a contraction between $\norm{z_{k+1} - \xs}^2$ and $\norm{z_k - \xs}^2$, we apply Lemma \ref{dis-contract} with $\mathcal{G}_y = \pf{y_k}, \mathcal{H}_y = \ph{y_k}$ and $z^+ = z_{k+1}$, which gives
\[
\innr{\ph{y_k}, z_k - \xs} = \frac{\alpha_k}{2} \left(\norm{z_k - \xs}^2 - \left(1 + \frac{\mu}{\alpha_k}\right)^2\norm{z_{k+1} - \xs}^2\right) + \frac{1}{2\alpha_k} \norm{\ph{y_k}}^2.
\]

Using this relation in \eqref{TM-h-contract}, expanding and re-arranging the terms, we conclude that
\[
\begin{aligned}
	&h(y_k) - \left(\frac{\tau^x_k - \mu\tau^z_k}{2\alpha_k} - \frac{1}{2(L - \mu)}\right)\norm{\ph{y_k}}^2 + \frac{\alpha_k(\tau^x_k - \mu\tau^z_k)}{2}\left(1 + \frac{\mu}{\alpha_k}\right)^2\norm{z_{k+1} - \xs}^2 \\ \leq{}& (1 - \tau^x_k)\left(h(y_{k-1}) - \frac{1}{2(L - \mu)} \norm{\ph{y_{k-1}}}^2\right)+ \frac{\alpha_k(\tau^x_k - \mu\tau^z_k)}{2} \norm{z_k - \xs}^2 \\
	&+ \left(\frac{1 - \tau^x_k}{L - \mu} - \tau^z_k\right) \innr{\ph{y_k}, \ph{y_{k-1}}}.
\end{aligned}
\]
It remains to impose parameter constraints according to the Lyapunov function.

\subsection{Proof of Proposition \ref{GTM-example}}
\label{app:GTM-example}
First, we can write the $k$th-update of G-TM with constant parameter as
\[
\begin{aligned}
	y_k &= (\tau_x - \tau_{z}\mu) z_k + \big(1 - (\tau_x - \tau_{z}\mu)\big) y_{k-1} - \tau_z\pf{y_{k-1}}, \\
	z_{k+1} &=  \frac{\alpha}{\alpha + \mu} z_k + \frac{\mu}{\alpha + \mu}y_k - \frac{1}{\alpha + \mu} \pf{y_k}.
\end{aligned}
\]

Substituting the constant parameter choice, we obtain
\[
\begin{aligned}
	y_k &= \frac{2}{\sqrt{\kappa} + 1} z_k + \frac{\sqrt{\kappa} - 1}{\sqrt{\kappa} + 1} \left(y_{k-1} - \frac{1}{L}\pf{y_{k-1}}\right), \\
	z_{k+1} &=  \left(1 - \frac{1}{\sqrt{\kappa}}\right) z_k + \frac{1}{\sqrt{\kappa}}y_k - \frac{1}{\sqrt{L\mu}} \pf{y_k}.
\end{aligned}
\]

For the objective function $f(x) = \frac{1}{2} \innr{\begin{bmatrix}
		L &0 \\
		0 &\mu 
	\end{bmatrix}x,x}$, the update can be further expanded as
\[
\begin{aligned}
	y_k &= \frac{2}{\sqrt{\kappa} + 1} z_k + \begin{bmatrix}
		0 &0 \\
		0 &\frac{(\sqrt{\kappa} - 1)^2}{\kappa} 
	\end{bmatrix} y_{k-1}, \\
	z_{k+1} &=  \left(1 - \frac{1}{\sqrt{\kappa}}\right) z_k + \begin{bmatrix}
		-\frac{\kappa - 1}{\sqrt{\kappa}} &0 \\
		0 &0
	\end{bmatrix} y_k.
\end{aligned}
\]

Thus,
\[
z_{k+1} = \left(1 - \frac{1}{\sqrt{\kappa}}\right) \begin{bmatrix}
	-1 &0 \\
	0 &1
\end{bmatrix} z_k \Longrightarrow \norm{z_{k+1} - \xs}^2 =  \left(1 - \frac{1}{\sqrt{\kappa}}\right)^2 \norm{z_k - \xs}^2,
\]
as desired.

\section{Proofs for Section \ref{sec:variance-reduction}}
\label{app:proofs_of_vr}

\subsection{Proof of Theorem \ref{TM-SVRG-contract}}
\label{app:proof_BS_SVRG}
For simplicity of presentation, we omit the superscript $s$ for iterates in the same epoch.

Using the trick in Lemma \ref{func-contract} for the recursion $y_{k} = \tau_x z_k + \left(1 - \tau_x\right) \tilde{x}_s +  \tau_{z}\left(\mu(\tilde{x}_s - z_k) - \pf{\tilde{x}_s}\right)$ and strengthening the convexity arguments by interpolation condition, we obtain
\[
\begin{aligned}
	h(y_k) \leq{}& \frac{1 - \tau_x}{\tau_x}\innr{\ph{y_k}, \tilde{x}_s - y_k} + \frac{\tau_{z}}{\tau_x} \innr{\ph{y_k}, \mu(\tilde{x}_s - z_k) - \pf{\tilde{x}_s}}+ \innr{\ph{y_k}, z_k - \xs} \\
	&- \frac{1}{2(L-\mu)} \norm{\ph{y_k}}^2.
\end{aligned}
\]
Note that here the inner product $\innr{\ph{y_k}, \tilde{x}_s - y_k}$ is not upper bounded as before. This term is preserved to deal with the variance. 

By the definition of $h$, $\mu(\tilde{x}_s - z_k) - \pf{\tilde{x}_s} = \mu(\xs - z_k) - \ph{\tilde{x}_s}$. Applying Lemma \ref{dis-contract} with $\mathcal{H}_y = \mathcal{H}^{\text{SVRG}}_{y_k}, \mathcal{G}_y = \mathcal{G}^{\text{SVRG}}_{y_k}, z^+ = z_{k+1}$ and taking the expectation, we can conclude that
\[
\begin{aligned}
	h(y_k) \leq{}& \frac{1 - \tau_x}{\tau_x}\innr{\ph{y_k}, \tilde{x}_s - y_k} - \frac{\tau_{z}}{\tau_x} \innr{\ph{y_k}, \ph{\tilde{x}_s}} - \frac{1}{2(L-\mu)} \norm{\ph{y_k}}^2 \\
	&+ \left(1 - \frac{\mu\tau_z}{\tau_x}\right)\frac{\alpha}{2} \left(\norm{z_k - \xs}^2 - \left(1 + \frac{\mu}{\alpha}\right)^2\Eik{\norm{z_{k+1} - \xs}^2}\right) \\&+ \left(\frac{1}{2\alpha} - \frac{\mu\tau_z}{2\alpha\tau_x}\right) \Eik{\norm{\mathcal{H}^{\text{SVRG}}_{y_k}}^2}.
\end{aligned}
\]

To bound the shifted moment, we apply the interpolation condition of $h_{i_k}$, i.e.,
\[
\begin{aligned}
	\Eik{\norm{\mathcal{H}^{\text{SVRG}}_{y_k}}^2} ={}& \Eik{\norm{\phik{y_k} - \phik{\tilde{x}_s}}^2} + 2\innr{\ph{y_k}, \ph{\tilde{x}_s}} - \norm{\ph{\tilde{x}_s}}^2\\
	\leq{}& 2(L-\mu) \big(h(\tilde{x}_s) - h(y_k) - \innr{\ph{y_k}, \tilde{x}_s - y_k}\big) + 2\innr{\ph{y_k}, \ph{\tilde{x}_s}}\\ 
	&- \norm{\ph{\tilde{x}_s}}^2.
\end{aligned}
\]


After re-arranging the terms, we obtain
\[
\begin{aligned}
	h(y_k) \leq{}& \left(1 - \frac{\mu\tau_z}{\tau_x}\right) \frac{L-\mu}{\alpha} \big(h(\tilde{x}_s) - h(y_k)\big) \\&+\left[\frac{1 - \tau_x}{\tau_x} - \left(1 - \frac{\mu\tau_z}{\tau_x}\right) \frac{L-\mu}{\alpha}\right]\innr{\ph{y_k}, \tilde{x}_s - y_k} \\
	&+ \left(1 - \frac{\mu\tau_z}{\tau_x}\right)\frac{\alpha}{2} \left(\norm{z_k - \xs}^2 - \left(1 + \frac{\mu}{\alpha}\right)^2\Eik{\norm{z_{k+1} - \xs}^2}\right) \\
	&+ \left(\frac{1}{\alpha} - \frac{\mu\tau_z}{\alpha\tau_x} - \frac{\tau_{z}}{\tau_x}\right) \innr{\ph{y_k}, \ph{\tilde{x}_s}}
	- \frac{1}{2(L-\mu)}\norm{\ph{y_k}}^2 \\
	&- \left(\frac{1}{2\alpha} - \frac{\mu\tau_z}{2\alpha\tau_x}\right)\norm{\ph{\tilde{x}_s}}^2.
\end{aligned}
\]

To cancel $\innr{\ph{y_k}, \tilde{x}_s - y_k}$, we choose $\tau_z$ such that $\frac{1 - \tau_x}{\tau_x} = \left(1 - \frac{\mu\tau_z}{\tau_x}\right) \frac{L-\mu}{\alpha}$, which gives
\begin{equation}\label{BS_SVRG_n1}
	\begin{aligned}
		h(y_k) \leq{}& (1 - \tau_x) h(\tilde{x}_s) + \frac{\alpha^2 (1 - \tau_x)}{2(L - \mu)} \left(\norm{z_k - \xs}^2 - \left(1 + \frac{\mu}{\alpha}\right)^2\Eik{\norm{z_{k+1} - \xs}^2}\right) \\
		&+ \frac{\alpha + \mu - (\alpha + L)\tau_x}{(L - \mu)\mu} \innr{\ph{y_k}, \ph{\tilde{x}_s}}
		- \frac{\tau_x}{2(L-\mu)}\norm{\ph{y_k}}^2 \\
		&- \frac{1 - \tau_x}{2(L-\mu)}\norm{\ph{\tilde{x}_s}}^2.
	\end{aligned}
\end{equation}

In view of the Lyapunov function $T_s \triangleq  h(\tilde{x}_s) - c_1 \norm{\ph{\tilde{x}_s}}^2 + \frac{\lambda}{2} \norm{z^s_0 - \xs}^2$, there are two ways to deal with the inner product $\innr{\ph{y_k}, \ph{\tilde{x}_s}}$:

{\bf\noindent Case I ($c_1 = 0$):} Choosing $\tau_x$ such that $\alpha + \mu - (\alpha + L)\tau_x = 0 \Longrightarrow \tau_x = \frac{\alpha + \mu}{\alpha + L}$ and dropping the negative gradient norms in \eqref{BS_SVRG_n1}, we arrive at \eqref{BS_SVRG_n2} with $c_1 = 0$.

{\bf\noindent Case II ($c_1 \neq 0$):} Denoting $\gamma = \frac{\abs{\alpha + \mu - (\alpha + L)\tau_x}}{(L - \mu)\mu}$ and using Young's inequality for $\innr{\ph{y_k}, \ph{\tilde{x}_s}}$ with parameter $\beta > 0$, we can bound \eqref{BS_SVRG_n1} as
\begin{equation}\label{TM-SVRG-P1}
	\begin{aligned}
		h(y_k)\leq{}& (1 - \tau_x) h(\tilde{x}_s)+ \frac{\alpha^2 (1 - \tau_x)}{2(L - \mu)} \left(\norm{z_k - \xs}^2 - \left(1 + \frac{\mu}{\alpha}\right)^2\Eik{\norm{z_{k+1} - \xs}^2}\right) \\&+ \left(\frac{\beta\gamma}{2} - \frac{\tau_x}{2(L-\mu)}\right)\norm{\ph{y_k}}^2 - \left(\frac{1 - \tau_x}{2(L - \mu)} -  \frac{\gamma}{2\beta}\right)\norm{\ph{\tilde{x}_s}}^2.
	\end{aligned}
\end{equation}

We require $\gamma \neq 0$ and choose $\beta > 0$ such that
\[
\frac{\beta\gamma}{2} - \frac{\tau_x}{2(L-\mu)} = \frac{1}{1 - \tau_x} \left(\frac{1 - \tau_x}{2(L - \mu)} -  \frac{\gamma}{2\beta}\right) = c_1 >0.
\]

It can be verified that this requirement and the existence of $\beta$ are equivalent to the following constraints:
\[
\begin{cases}
	\tau_x \neq \frac{\alpha + \mu}{\alpha + L},\\
	(1 + \tau_x)^2(1 - \tau_x)\geq 4\left(\left(\frac{\alpha}{\mu} + 1\right) - \left(\frac{\alpha}{\mu} + \kappa\right)\tau_x\right)^2.
\end{cases}
\]

Under these constraints, denoting  $\Delta = \frac{(1 + \tau_x)^2}{(L - \mu)^2} - \frac{4\gamma^2}{1 - \tau_x} \geq 0$, we can choose $\beta = \frac{1 + \tau_x}{2\gamma(L - \mu)} + \frac{\sqrt{\Delta}}{2\gamma}$, which ensures $c_1 \in \left]0, \frac{1}{2(L - \mu)} \right[$. 

Let $c_2 \triangleq \frac{\alpha^2 (1 - \tau_x)}{L - \mu}$. These two cases result in the same inequality:
\begin{equation}\label{BS_SVRG_n2}
	\begin{aligned}
		h(y_k) -  c_1\norm{\ph{y_k}}^2 \leq{}& (1 - \tau_x) \big(h(\tilde{x}_s) - c_1\norm{\ph{\tilde{x}_s}}^2\big) 
		\\&+ \frac{c_2}{2} \left(\norm{z_k - \xs}^2 - \left(1 + \frac{\mu}{\alpha}\right)^2\Eik{\norm{z_{k+1} - \xs}^2}\right).
	\end{aligned}
\end{equation}

Finally, summing the above inequality from $k=0,\ldots, m-1$ with weight $\left(1 + \frac{\mu}{\alpha}\right)^{2k}$, we conclude that
\begin{equation}\label{TM-SVRG-final}
	\begin{aligned}
		&\E{h(\tilde{x}_{s+1}) -  c_1\norm{\ph{\tilde{x}_{s+1}}}^2} =\sum_{k=0}^{m-1} {\frac{1}{\widetilde{\omega}}\left(1 + \frac{\mu}{\alpha}\right)^{2k}\E{h(y^s_k) -  c_1\norm{\ph{y^s_k}}^2}}\\ \leq{}& (1 - \tau_x) \big(h(\tilde{x}_s)- c_1\norm{\ph{\tilde{x}_s}}^2\big) + \frac{c_2}{2\widetilde{\omega}} \left(\norm{z^{s}_0 - \xs}^2 - \left(1 + \frac{\mu}{\alpha}\right)^{2m}\E{\norm{z^s_{m} - \xs}^2}\right).
	\end{aligned}
\end{equation}

Imposing the constraint $\left(1 + \frac{\mu}{\alpha}\right)^{2m} (1 - \tau_x) \leq 1$ completes the proof.

\subsection{Proof of Proposition \ref{TM-SVRG-choice}}
\label{app:TM-SVRG-choice}
The choice 
\[
\begin{cases}
	\alpha = \sqrt{cm\mu L} - \mu, \\
	\tau_x = \left(1 - \frac{1}{c\kappa}\right)\frac{\alpha + \mu}{\alpha + L} = \left(1 - \frac{1}{c\kappa}\right)\frac{\sqrt{cm\kappa}}{\sqrt{cm\kappa} + \kappa - 1},
\end{cases}
\]
and the constraints 
\begin{gather}
	(1 + \tau_x)^2(1 - \tau_x)\geq 4\left(\left(\frac{\alpha}{\mu} + 1\right) - \left(\frac{\alpha}{\mu} + \kappa\right)\tau_x\right)^2,\label{C2} \\
	\left(1 + \frac{\mu}{\alpha}\right)^{2m} (1 - \tau_x) \leq 1,\label{C1}
\end{gather}
are put here for reference.

Note that for $m\in \left(0, \frac{3}{4}\kappa \right]$,
$\tau_x = \frac{c\kappa - 1}{c\kappa + \sqrt{\frac{c\kappa}{m}}(\kappa - 1)}$
increases monotonically  and $\frac{1 + \tau_x}{m}$ decreases monotonically as $m$ increases.
Thus, for the constraint \eqref{C2}, letting
\[
\phi(m, \kappa) \triangleq \frac{( 1 + \tau_x)^2(1 - \tau_x)}{\left(\left(\frac{\alpha}{\mu} + 1\right) - \left(\frac{\alpha}{\mu} + \kappa\right)\tau_x\right)^2} = \frac{1 + \tau_x}{m}\left(1 - \tau_x^2\right)c\kappa,
\]
we have $\phi(m, \kappa)$ decreases monotonically as $m$ increases.

When $m = \frac{3}{4} \kappa$, $\tau_x = \frac{c\kappa - 1}{\left(c + \sqrt{\frac{4c}{3}}\right)\kappa - \sqrt{\frac{4c}{3}}}$. For $\kappa \geq 1$, if $c + \sqrt{\frac{4c}{3}} - c\sqrt{\frac{4c}{3}} \leq 0 \Leftrightarrow c \geq \frac{(\sqrt{3} + \sqrt{19})^2}{16} \approx 2.319$, we have $\tau_x$ decreases monotonically as $\kappa$ increases. In this case, letting $\kappa \rightarrow \infty$, we conclude that $\tau_x > \frac{c}{c + \sqrt{\frac{4c}{3}}} > \frac{1}{3}$, which implies that $(1+\tau_x)^2(1 - \tau_x)$ increases monotonically as $\tau_x$ decreases. Thus,
\[
\phi(m, \kappa) \geq \phi\left(\frac{3}{4}\kappa, \kappa\right) \geq \phi\left(\frac{3}{4}, 1\right) = \frac{4}{3}\left(1 + \frac{c-1}{c}\right)\left(1 - \left(\frac{c-1}{c}\right)^2\right)c.
\]
To meet the constraint \eqref{C2}, we require $c \geq 2 + \sqrt{3}\approx 3.74$.

For constraint \eqref{C1}, defining
\[
\psi(m, \kappa) \triangleq \left(\frac{\alpha + \mu}{\alpha}\right)^{2m} (1 - \tau_x) = \left(1 + \frac{1}{\sqrt{cm\kappa}-1}\right)^{2m} \frac{\sqrt{cm\kappa} + c\kappa(\kappa - 1)}{(\sqrt{cm\kappa} - 1 + \kappa)c\kappa},
\]
we have $\frac{\partial \psi}{\partial m} =$
\[
\begin{aligned}
	\left(1 + \frac{1}{\sqrt{cm\kappa}-1}\right)^{2m}\Bigg[&\left(2\ln{\left(1 + \frac{1}{\sqrt{cm\kappa} - 1}\right)- \frac{1}{\sqrt{cm\kappa} - 1}}\right) \frac{\sqrt{cm\kappa} + c\kappa(\kappa - 1)}{(\sqrt{cm\kappa} - 1 + \kappa)c\kappa} \\
	&- \frac{(\kappa-1)(c\kappa - 1)}{2\sqrt{cm\kappa}\big(\sqrt{cm\kappa} - 1 + \kappa\big)^2}\Bigg].
\end{aligned}
\]				

Denote $q = \sqrt{cm\kappa} - 1>0$. The roots of $\frac{\partial\psi}{\partial m}$ are identified by the following equation:		
\[				
s(q) \triangleq 2 \ln{\left(1 + \frac{1}{q}\right)} - \frac{1}{q} - \frac{b_0}{(q+1)(q+ \kappa)(q + b_1)} = 0,
\]
where $b_0=\frac{c\kappa}{2}(\kappa-1)(c\kappa - 1), b_1=1 + c\kappa(\kappa - 1)$. Taking derivative, we see that when $q\rightarrow 0$, $s'(q) \geq \frac{1}{q^2} - \frac{2}{q(1 + q)}\rightarrow \infty$. We can arrange the equation $s'(q) = 0$ as finding the real roots of a polynomial. By Descartes' rule of signs, this equation has exactly one positive root (with $c \geq 2+\sqrt{3}$, we have $\kappa b_1 - 1 - b_0 \leq 0$ for any $\kappa \geq 1$ and then there is exactly one sign change in the polynomial). Thus, as $q$ increases, $s(q)$ first increases monotonically to the unique root and then decreases monotonically.

To see that $s(q)$ has exactly one root, let $q \rightarrow 0, s(q) \leq 2 \ln{\left(1 + \frac{1}{q}\right)} - \frac{1}{q} \rightarrow -\infty$; when $q$ is large enough (e.g., $q>2$ and $(q+\kappa)(q+b_1) > 2b_0$), $s(q)>0$; let $q \rightarrow \infty, s(q) \rightarrow 0$. These facts suggest that $s(q)$ has a unique root. Thus, we conclude that, as $m$ increases, $\psi(m, \kappa)$ first decreases monotonically to the unique root and then increases monotonically, which means that for $m\in [2, \frac{3}{4}\kappa], \psi(m, \kappa) \leq \max{\left\{\psi(2, \kappa), \psi\left(\frac{3}{4}\kappa, \kappa\right)\right\}}$. 

For $\psi(2, \kappa)$, $\psi'(2, \kappa) = \left(1 + \frac{1}{\sqrt{2c\kappa} - 1}\right)^4 \left(\sqrt{2c\kappa}+\kappa-1\right)^{-2}\left(\sqrt{2c\kappa}-1\right)^{-1} \ell(\kappa)$, where $\ell(\kappa)$ is a polynomial:
\[
\ell(\kappa) \triangleq \left(c-2\right)\kappa-\frac{5\sqrt{2c}}{2}\kappa^{\frac{1}{2}} + (c + 1) - \left(\sqrt{\frac{c}{2}}+\frac{1}{\sqrt{2c}}\right)\kappa^{-\frac{1}{2}} - 3\kappa^{-1} + \frac{3}{\sqrt{2c}} \kappa^{-\frac{3}{2}}.
\]
It can be verified that with $c\geq 2+\sqrt{3}$, for any $\kappa \geq \frac{8}{3}, \ell'(\kappa) > 0$, which suggests that $\psi(2,\kappa) \leq \max{\left\{\psi\left(2, \frac{8}{3}\right), \psi(2, \infty)\right\}} \leq 1$ (with $c \geq 2+\sqrt{3}, \psi\left(2, \frac{8}{3}\right) \leq 0.953$ and $\psi\left(2, \infty\right) = 1$).

For $\psi\left(\frac{3}{4}\kappa, \kappa\right)$, $\psi'\left(\frac{3}{4}\kappa, \kappa\right) = \left(1 + \frac{2}{\sqrt{3c}\kappa - 2}\right)^{\frac{3}{2}\kappa}\left(\left(c+\sqrt{\frac{4c}{3}}\right)\kappa - \sqrt{\frac{4c}{3}}\right)^{-1} \omega_1 (\kappa)$, where 
\[
\omega_1(\kappa) \triangleq \left(\ln\left(1+\frac{2}{\sqrt{3c}\kappa-2}\right)-\frac{2}{\sqrt{3c}\kappa-2}\right)\left(\sqrt{3c}\kappa-\sqrt{3c}+\frac{3}{2}\right)\ +\frac{\sqrt{\frac{4c}{3}}c-c-\sqrt{\frac{4c}{3}}}{\left(c+\sqrt{\frac{4c}{3}}\right)\kappa-\sqrt{\frac{4c}{3}}}.
\]

Let $p=\sqrt{3c}\kappa - 2 > 0$, the roots of $\omega_1(\kappa)$ are determined by the equation
\[
\omega_2(p) \triangleq \ln\left(1+\frac{2}{p}\right)-\frac{2}{p}+\frac{\frac{3}{2+\sqrt{3c}}\left(\sqrt{\frac{4c}{3}}c-c-\sqrt{\frac{4c}{3}}\right)}{\left(p+\frac{4}{2+\sqrt{3c}}\right)\left(p+\frac{7}{2}-\sqrt{3c}\right)} = 0.
\]
To ensure that $\omega_2(p)$ increases monotonically as $p$ increases, it suffices to set $c \leq 3.817$ (which ensures that $\omega_2'(p) > 0$). Thus, for any $p>0$, $\omega_2(p) \leq \lim_{p\rightarrow \infty}{\omega_2(p)} = 0 \Rightarrow$ for any $\kappa \geq 1$, $\omega_1(\kappa) \leq 0$. Finally, we conclude that with $3.817\geq c \geq 2+\sqrt{3}$, $\psi\left(\frac{3}{4}\kappa, \kappa\right) \leq \psi\left(2, \frac{8}{3}\right) \leq 0.953$, which completes the proof.

\subsection{Proof of Proposition \ref{TM-SVRG-choice-well}}
\label{app:TM-SVRG-choice-well}

The choice $\begin{cases}
	\alpha = \frac{3L}{2}
	- \mu, \\
	\tau_x = \left(1 - \frac{1}{6m}\right)\frac{\alpha + \mu}{\alpha + L} = \left(1 - \frac{1}{6m}\right) \frac{3\kappa}{5\kappa - 2},
\end{cases}$ is put here for reference.

We examine the constraint $(1 + \tau_x)^2(1 - \tau_x)\geq 4\left(\left(\frac{\alpha}{\mu} + 1\right) - \left(\frac{\alpha}{\mu} + \kappa\right)\tau_x\right)^2$. Let
\[
\phi(m, \kappa) \triangleq \frac{(1 + \tau_x)^2(1 - \tau_x)}{4\left(\left(\frac{\alpha}{\mu} + 1\right) - \left(\frac{\alpha}{\mu} + \kappa\right)\tau_x\right)^2} = \frac{(1 + \tau_x)^2(1 - \tau_x)4m^2}{\kappa^2}.
\]

For $m \geq \frac{3}{4}\kappa$, we have $\tau_x$ and $(1 - \tau_x)m$ increases monotonically as $m$ increases. Thus, $\phi(m, \kappa)$ increases as $m$ increases $\Longrightarrow \phi(m, \kappa) \geq \phi(\frac{3}{4}\kappa, \kappa)$.

$\phi(\frac{3}{4}\kappa, \kappa) = \frac{9}{4}(1 + \tau_x)^2(1 - \tau_x)$ and $\tau_x = \frac{9\kappa - 2}{15\kappa - 6}$ in this case. Note that for $\kappa \geq 1$, $\tau_x$ decreases as $\kappa$ increases and let $\kappa \rightarrow \infty$, we conclude that $\tau_x > \frac{3}{5} > \frac{1}{3} \Longrightarrow (1 + \tau_x)^2(1 - \tau_x)$ increases as $\tau_x$ decreases. Thus, $\phi(\frac{3}{4}\kappa, \kappa) \geq \phi(\frac{3}{4}, 1) > 1$, the constraint is satisfied.

Using this choice, we can write the per-epoch contraction \eqref{TM-SVRG-final} in Theorem \ref{TM-SVRG-contract} as 
\[
\begin{aligned}
	&\E{h(\tilde{x}_{s+1}) -  c_1\norm{\ph{\tilde{x}_{s+1}}}^2} + \frac{\alpha^2 (1 - \tau_x)}{2\widetilde{\omega}(L - \mu)}\left(1 + \frac{\mu}{\alpha}\right)^{2m}\E{\norm{z^{s+1}_{0} - \xs}^2} \\ \leq{}& (1 - \tau_x) \big(h(\tilde{x}_s)- c_1\norm{\ph{\tilde{x}_s}}^2\big) + \frac{\alpha^2 (1 - \tau_x)}{2\widetilde{\omega}(L - \mu)} \norm{z^{s}_0 - \xs}^2.
\end{aligned}
\]

Note that for $\frac{m}{\kappa} > \frac{3}{4}$, $\tau_x > \frac{1}{2}$ and by Bernoulli's inequality, $\left(1 + \frac{\mu}{\alpha}\right)^{2m} \geq 1 + \frac{2m\mu}{\alpha} = 1 + \frac{4m}{3\kappa - 2} > 2$. Let $\lambda = \frac{2\alpha^2 (1 - \tau_x)}{\widetilde{\omega}(L - \mu)}$. The above contraction becomes
\[
\begin{aligned}
	&\E{h(\tilde{x}_{s+1}) -  c_1\norm{\ph{\tilde{x}_{s+1}}}^2} + \frac{\lambda}{2}\E{\norm{z^{s+1}_{0} - \xs}^2}  \\
	\leq{}& \frac{1}{2} \cdot  \left(h(\tilde{x}_s)- c_1\norm{\ph{\tilde{x}_s}}^2 + \frac{\lambda}{2} \norm{z^{s}_0 - \xs}^2\right).
\end{aligned}
\]

Telescoping this inequality from $S-1$ to $0$, we obtain $T_S \leq \frac{1}{2^S} T_0$, and since $m=2n$, these imply an $O(n\log{\frac{1}{\epsilon}})$ iteration complexity.

\subsection{BS-SAGA}
\label{app:BS-SAGA}

\begin{algorithm}[t]
	\caption{SAGA Boosted by Shifting objective (BS-SAGA)}
	\label{TM-SAGA}
	\renewcommand{\algorithmicrequire}{\textbf{Input:}}
	\renewcommand{\algorithmicensure}{\textbf{Initialize:}}
	\begin{algorithmic}[1]
		\REQUIRE Parameters $\alpha > 0, \tau_x \in ]0, 1[$ and initial guess $x_0 \in \R^d$, iteration number $K$.
		\ENSURE $z_0 = x_0, \tau_z = \frac{\tau_x}{\mu} - \frac{\alpha(1 - \tau_x)}{\mu(L-\mu)}$, a point table $\phi^0 \in \R^{d\times n}$ with $\forall i\in [n], \phi^0_i = x_0$, running averages for the point table and its gradients.
		\FOR{$k=0, \ldots, K-1$}
		\STATE Sample $i_k$ uniformly in $[n]$, set $\phi_{i_k}^{k+1} = \tau_x z_k + \left(1 - \tau_x\right) \phi_{i_k}^{k} +  \tau_{z}\big(\mu (\bar{\phi}^k - z_k) - \frac{1}{n} \sum_{i=1}^n {\pfi{\phi^k_i}} \big)$ and keep other entries unchanged (i.e., for $i\neq i_k, \phi^{k+1}_i = \phi^k_i$).
		\STATE $z_{k+1} = \arg\min_{x} \Big\{\big\langle{\mathcal{G}^{\text{SAGA}}_{\phi_{i_k}^{k+1}}, x}\big\rangle + (\alpha/2) \norm{x - z_k}^2 + (\mu/2) \norm{x - \phi_{i_k}^{k+1}}^2 \Big\}$.
		\STATE Update the running averages according to the change in $\phi^{k+1}$.
		\ENDFOR
		\renewcommand{\algorithmicensure}{\textbf{Output:}}
		\ENSURE $z_K$.
	\end{algorithmic}
\end{algorithm}
To make the notations specific, we define 
\[
\begin{aligned}
	&\mathcal{H}^{\text{SAGA}}_{x_k} \triangleq \phik{x_k} - \phik{\phi^k_{i_k}} + \frac{1}{n}\sum_{i=1}^n{\phii{\phi^k_i}}\\
	\Rightarrow{}&\mathcal{G}^{\text{SAGA}}_{x_k} \triangleq \pfik{x_k} - \pfik{\phi^k_{i_k}} + \frac{1}{n}\sum_{i=1}^n{\pfi{\phi^k_i}}- \mu\left(\bar{\phi}^k - \phi_{i_k}^k\right),
\end{aligned}
\]
where $\phi^k \in \R^{d\times n}$ is a point table that stores $n$ previously chosen random anchor points and $\bar{\phi}^k \triangleq \frac{1}{n} \sum_{i = 1}^n{\phi^k_i}$ denotes the average of point table.

The Lyapunov function (with $c_1 \in \left[0, \frac{1}{2(L-\mu)}\right], \lambda > 0$) is put here for reference:
\begin{equation}\label{TM-SAGA-Lya}
	T_k = \frac{1}{n}\sum_{i=1}^n{h_{i}(\phi^{k}_{i})} - c_1\norml{\frac{1}{n}\sum_{i=1}^n{\phii{\phi^k_{i}}}}^2 + \frac{\lambda}{2} \norm{z_k - \xs}^2.
\end{equation}

We present the SAGA variant in Algorithm \ref{TM-SAGA}. In the following theorem, we only consider a simple case with $c_1 = 0$ in $T_k$. It is possible to analyze BS-SAGA with $c_1 \neq 0$ as is the case for BS-SVRG (the analysis in Appendix \ref{app:proof_BS_SVRG}). However, it leads to highly complicated parameter constraints. We provide a simple parameter choice similar to the one in Proposition \ref{TM-SVRG-choice-numerical}.

\begin{app-theorems}\label{TM-SAGA-contract} In Algorithm~\ref{TM-SAGA}, if we choose $\alpha, \tau_x$ as
	\begin{equation}\label{TM-SAGA-choice}
		\begin{cases}
			\textup{$\alpha$ is solved from the equation } \left(1 + \frac{ \mu}{\alpha}\right)^2\left(1 - \frac{\alpha + \mu}{(\alpha + L)n}\right) = 1,\\
			\tau_x = \frac{\alpha + \mu}{\alpha + L},
		\end{cases}
	\end{equation}
	the following per-iteration contraction holds for the Lyapunov function defined at \eqref{TM-SAGA-Lya} (with $c_1 = 0$).
	\[
	\text{With } \lambda = \frac{(1 - \tau_x)\left(\alpha + \mu\right)^2}{(L - \mu)n} ,\ \
	\Eik{T_{k+1}} \leq \left(1 + \frac{\mu}{\alpha}\right)^{-2} T_k, \text{ for } k\geq 0.
	\]
\end{app-theorems}

Regrading the rate, from \eqref{TM-SAGA-choice}, we can figure out that $\alpha$ is the unique positive root of the cubic equation:
\[
\left(\frac{\alpha}{\mu}\right)^3 - (2n-3)\left(\frac{\alpha}{\mu}\right)^2 - (2n\kappa + n - 3)\left(\frac{\alpha}{\mu}\right) - (n\kappa - 1) = 0.
\]
Using a similar argument as in Theorem \ref{TM-Point-SAGA-contract}, we can show that $\frac{\alpha}{\mu} = O(n + \sqrt{n\kappa})$, and thus conclude an $O\big((n + \sqrt{n\kappa})\log{\frac{1}{\epsilon}}\big)$ expected  complexity for BS-SAGA. Interestingly, this rate is always slightly slower than that of BS-Point-SAGA.

\subsubsection{Proof of Theorem \ref{TM-SAGA-contract}}
To simplify the notations in this proof, we let $\Phi^k \triangleq \frac{1}{n} \sum_{i=1}^n {h_i(\phi^k_i)}$ and $\nabla\Phi^k \triangleq \frac{1}{n} \sum_{i=1}^n {\phii{\phi^k_i}}$.

Using the trick in Lemma \ref{func-contract} (with $f = h_{i_k}$) for $\phi^{k+1}_{i_k}$, strengthening the convexity with the interpolation condition and taking the expectation, we obtain
\[
\begin{aligned}
	\Eik{h_{i_k}(\phi^{k+1}_{i_k}) } \leq{}& \frac{1 - \tau_x}{\tau_x}\Eik{\innr{\phik{\phi^{k+1}_{i_k}}, \phi^{k}_{i_k} - \phi^{k+1}_{i_k}}}+ \Eik{\innr{\phik{\phi^{k+1}_{i_k}}, z_k - \xs}}  \\
	&+ \frac{\tau_{z}}{\tau_x} \Eik{\innr{\phik{\phi^{k+1}_{i_k}}, \mu (\bar{\phi}^k - z_k) - \frac{1}{n} \sum_{i=1}^n {\pfi{\phi^k_i}}}} \\
	&- \frac{1}{2(L - \mu)} \Eik{\norml{\phik{\phi^{k+1}_{i_k}}}^2}.
\end{aligned}
\]

Note that by the definition of $h_i$, $\mu (\bar{\phi}^k - z_k) - \frac{1}{n} \sum_{i=1}^n {\pfi{\phi^k_i}} = \mu (\xs - z_k) - \nabla\Phi^k$, and thus
\begin{equation}\label{TM-SAGA-P5}
	\begin{aligned}
		\Eik{h_{i_k}(\phi^{k+1}_{i_k}) } \leq{}& \frac{1 - \tau_x}{\tau_x}\Eik{\innr{\phik{\phi^{k+1}_{i_k}}, \phi^{k}_{i_k} - \phi^{k+1}_{i_k}}} - \frac{\tau_{z}}{\tau_x} \Eik{\innr{\phik{\phi^{k+1}_{i_k}}, \nabla\Phi^k}} \\
		&+ \left(1 - \frac{\mu\tau_{z}}{\tau_x}\right)\Eik{\innr{\phik{\phi^{k+1}_{i_k}}, z_k - \xs}} \\
		&- \frac{1}{2(L - \mu)} \norml{\Eik{\phik{\phi^{k+1}_{i_k}}}}^2,
	\end{aligned}
\end{equation}
which also uses Jensen's inequality, i.e., $\Eik{\norm{\phik{\phi^{k+1}_{i_k}}}^2} \geq\norm{\Eik{\phik{\phi^{k+1}_{i_k}}}}^2$.

Using Lemma \ref{dis-contract} with $\mathcal{H}_y = \mathcal{H}^{\text{SAGA}}_{\phi^{k+1}_{i_k}}, \mathcal{G}_y = \mathcal{G}^{\text{SAGA}}_{\phi_{i_k}^{k+1}}, z^+ = z_{k+1}$ and taking the expectation, we obtain
\begin{equation}\label{TM-SAGA-P6}
	\begin{aligned}
		\Eik{\innr{\phik{\phi^{k+1}_{i_k}}, z_k - \xs}} ={}& \frac{\alpha}{2} \left(\norm{z_k - \xs}^2 - \left(1 + \frac{\mu}{\alpha}\right)^2\Eik{\norm{z_{k+1} - \xs}^2}\right) \\
		&+ \frac{1}{2\alpha} \Eik{\norml{\mathcal{H}^{\text{SAGA}}_{\phi^{k+1}_{i_k}}}^2}.
	\end{aligned}
\end{equation}

Using the interpolation condition of $h_{i_k}$ to bound the stochastic moment,
\begin{align}
	\Eik{\norml{\mathcal{H}^{\text{SAGA}}_{\phi^{k+1}_{i_k}}}^2} ={}& \Eik{\norm{\phik{\phi^{k+1}_{i_k}} - \phik{\phi^k_{i_k}}}^2} + 2\Eik{\innr{\phik{\phi^{k+1}_{i_k}}, \nabla \Phi^k}}\nonumber \\&- \norm{\nabla \Phi^k}^2 \nonumber\\
	\leq{}&2(L-\mu)\big(\Phi^k - \Eik{h_{i_k}(\phi^{k+1}_{i_k})} - \Eik{\innr{\phik{\phi^{k+1}_{i_k}}, \phi^k_{i_k} - \phi^{k+1}_{i_k}}}\big) \label{TM-SAGA-P7}\\&+ 2\Eik{\innr{\phik{\phi^{k+1}_{i_k}}, \nabla \Phi^k}} - \norm{\nabla \Phi^k}^2.\nonumber
\end{align}

Based on the updating rules of $\phi^{k+1}$, the following relations hold 
\begin{gather}
	\Eik{\Phi^{k+1}} = \frac{1}{n}\Eik{h_{i_k}(\phi_{i_k}^{k+1})} +  \frac{n-1}{n}\Phi^k, \label{TM-SAGA-P1}\\
	\Eik{\nabla \Phi^{k+1}} =  \frac{1}{n}\Eik{\phik{\phi^{k+1}_{i_k}}} + \frac{n-1}{n} \nabla\Phi^k,\label{TM-SAGA-P2}
\end{gather}
where \eqref{TM-SAGA-P2} implies that
\begin{align}
	&\begin{aligned}\normb{\Eik{\phik{\phi^{k+1}_{i_k}}}}^2
		={}& n^2\norm{\Eik{\nabla \Phi^{k+1}}}^2 - 2(n^2-n)\innr{\Eik{\nabla \Phi^{k+1}}, \nabla\Phi^k} \\&+ (n-1)^2\norm{\nabla\Phi^k}^2, 
	\end{aligned}\label{TM-SAGA-P3}\\
	&\Eik{\innr{\phik{\phi^{k+1}_{i_k}}, \nabla\Phi^k}} ={} n\innr{\Eik{\nabla \Phi^{k+1}}, \nabla\Phi^k} - (n-1) \norm{\nabla \Phi^k}^2.\label{TM-SAGA-P4}
\end{align}

Then, expanding \eqref{TM-SAGA-P5} using \eqref{TM-SAGA-P6}, \eqref{TM-SAGA-P7}, \eqref{TM-SAGA-P3} and \eqref{TM-SAGA-P4}, we obtain
\[
\begin{aligned}
	\frac{1}{n}\Eik{h_{i_k}(\phi^{k+1}_{i_k}) } \leq{}& \left[\frac{1 - \tau_x}{\tau_x n} - \left(1 - \frac{\mu\tau_{z}}{\tau_x}\right)\frac{L-\mu}{\alpha n}\right]\Eik{\innr{\phik{\phi^{k+1}_{i_k}}, \phi^{k}_{i_k} - \phi^{k+1}_{i_k}}} \\
	&+ \left(1 - \frac{\mu\tau_{z}}{\tau_x}\right)\frac{L-\mu}{\alpha n} \Big(\Phi^k - \Eik{h_{i_k}(\phi^{k+1}_{i_k})}\Big) \\
	&+ \left(1 - \frac{\mu\tau_{z}}{\tau_x}\right)\frac{\alpha}{2n} \left(\norm{z_k - \xs}^2 - \left(1 + \frac{\mu}{\alpha}\right)^2\Eik{\norm{z_{k+1} - \xs}^2}\right) \\
	&+ \left[\frac{1}{\alpha} - \frac{\mu\tau_{z}}{\alpha\tau_x} - \frac{\tau_{z}}{\tau_x}+\frac{n-1}{L - \mu}\right]\innr{\Eik{\nabla \Phi^{k+1}}, \nabla\Phi^k} \\
	&- \left[\frac{(n-1)^2}{2(L - \mu)n} + \left(1 - \frac{\mu\tau_{z}}{\tau_x}\right)\frac{1}{2\alpha n} + \left(\frac{1}{\alpha} - \frac{\mu\tau_{z}}{\alpha\tau_x} - \frac{\tau_{z}}{\tau_x}\right)\frac{n-1}{n}\right]\norm{\nabla\Phi^k}^2\\
	&- \frac{n}{2(L - \mu)}\norm{\Eik{\nabla \Phi^{k+1}}}^2.
\end{aligned}
\]

Choosing $\tau_z$ such that $\frac{1 - \tau_x}{\tau_x} = \left(1 - \frac{\mu\tau_{z}}{\tau_x}\right)\frac{L-\mu}{\alpha}$, multiplying both sides by $\tau_x$ and using \eqref{TM-SAGA-P1}, we can simplify the above inequality as
\[
\begin{aligned}
	\Eik{\Phi^{k+1}} \leq{}& \left(1 - \frac{\tau_x}{n}\right) \Phi^k + \frac{\alpha^2(1 - \tau_x)}{2(L-\mu)n} \left(\norm{z_k - \xs}^2 - \left(1 + \frac{\mu}{\alpha}\right)^2\Eik{\norm{z_{k+1} - \xs}^2}\right) \\
	&+ \frac{\alpha + \mu - \tau_x (\alpha + L + \mu - \mu n)}{(L-\mu)\mu} \innr{\Eik{\nabla \Phi^{k+1}}, \nabla\Phi^k} \\
	&- \frac{(n-2)\tau_x + \frac{1}{n} + \left(\frac{\alpha}{\mu}+1 - 
		\left(\frac{\alpha}{\mu} + \kappa\right)\tau_x\right)\left(2-\frac{2}{n}\right)}{2(L-\mu)}\norm{\nabla\Phi^k}^2
	\\&- \frac{n\tau_x}{2(L - \mu)} \norm{\Eik{\nabla \Phi^{k+1}}}^2.
\end{aligned}
\]

Fixing $\tau_x = \frac{\alpha + \mu}{\alpha + L}$, we obtain
\[
\begin{aligned}
	\Eik{\Phi^{k+1}} \leq{}& \left(1 - \frac{\tau_x}{n}\right) \Phi^k + \frac{\alpha^2(1 - \tau_x)}{2(L-\mu)n} \left(\norm{z_k - \xs}^2 - \left(1 + \frac{\mu}{\alpha}\right)^2\Eik{\norm{z_{k+1} - \xs}^2}\right) \\
	&+ \frac{(n-1)\tau_x}{L-\mu} \innr{\Eik{\nabla \Phi^{k+1}}, \nabla\Phi^k} - \frac{n\tau_x}{2(L - \mu)} \norm{\Eik{\nabla \Phi^{k+1}}}^2\\
	&- \frac{(n-2)\tau_x + \frac{1}{n}}{2(L-\mu)}\norm{\nabla\Phi^k}^2.
\end{aligned}
\]

Using Young's inequality with $\beta > 0$, 
\[
\begin{aligned}
	\Eik{\Phi^{k+1}} \leq{}& \left(1 - \frac{\tau_x}{n}\right) \Phi^k + \frac{\alpha^2(1 - \tau_x)}{2(L-\mu)n} \left(\norm{z_k - \xs}^2 - \left(1 + \frac{\mu}{\alpha}\right)^2\Eik{\norm{z_{k+1} - \xs}^2}\right) \\
	&+ \frac{\beta(n-1)\tau_x - n\tau_x}{2(L-\mu)} \norm{\Eik{\nabla \Phi^{k+1}}}^2 + \frac{\frac{(n-1)\tau_x}{\beta}-(n-2)\tau_x - \frac{1}{n}}{2(L-\mu)}\norm{\nabla\Phi^k}^2.
\end{aligned}
\]

Let $\beta \in \left[\frac{n-1}{n-2 + \frac{1}{n\tau_x}}, \frac{n}{n-1}\right]$. The last two terms become non-positive, and thus we have
\[
\Eik{ \Phi^{k+1}} \leq{} \left(1 - \frac{\tau_x}{n}\right) \cdot \Phi^k + \frac{\alpha^2(1 - \tau_x)}{2(L-\mu)n} \left(\norm{z_k - \xs}^2 - \left(1 + \frac{\mu}{\alpha}\right)^2\Eik{\norm{z_{k+1} - \xs}^2}\right).
\]

Letting $\left(1 - \frac{\tau_x}{n}\right)\left(1 + \frac{\mu}{\alpha}\right)^2 = 1$ completes the proof.

\section{Proof for Section \ref{sec:proximal-point} (Theorem \ref{TM-Point-SAGA-contract})}
\label{app:proof_of_point}
Using Lemma \ref{point-contract} with the relations 
\[
\begin{gathered}
	x_{k+1} = \tprox^\alpha_{i_k} \left(x_k + \frac{1}{\alpha}\left(\pfik{\phi^k_{i_k}} - \frac{1}{n}\sum_{i=1}^n{\pfi{\phi^k_i}} + \mu \left(\frac{1}{n} \sum_{i = 1}^n{\phi^k_i} - \phi^k_{i_k}\right)\right)\right), \\
	\xs = \tprox^{\alpha}_{i_k} \left(\xs + \frac{1}{\alpha} \pfik{\xs}\right)\text{ and } \phi^{k+1}_{i_k} = x_{k+1},
\end{gathered}
\]
and based on that $\phii{x} = \pfi{x} - \pfi{\xs} - \mu(x - \xs)$, we have
\[
\begin{aligned}
	&\left(1 + \frac{2(\alpha + \mu)}{L - \mu}\right) \norm{\phik{\phi^{k+1}_{i_k}}}^2 + (\alpha + \mu)^2 \norm{x_{k+1} - \xs}^2 \\
	\leq{}& \alpha^2 \norml{x_k - \xs + \frac{1}{\alpha} \left(\phik{\phi^k_{i_k}} - \frac{1}{n}\sum_{i=1}^n{\phii{\phi^k_i}}\right)}^2.
\end{aligned}
\]

Expanding the right side, taking the expectation and using $\E{\norm{X - \mathbb{E}X}^2} \leq \E{\norm{X}^2}$, we obtain
\[
\begin{aligned}
	&\left(1 + \frac{2(\alpha + \mu)}{L - \mu}\right) \Eik{\norm{\phik{\phi^{k+1}_{i_k}}}^2} + (\alpha + \mu)^2 \Eik{\norm{x_{k+1} - \xs}^2}
	\\ \leq{}& \alpha^2 \norm{x_k - \xs}^2 + \frac{1}{n}\sum_{i=1}^n{\norm{\phii{\phi^k_{i}}}^2}.
\end{aligned}
\]

Note that by construction,
\[
\Eik{\sum_{i=1}^n{\norm{\phii{\phi^{k+1}_{i}}}^2}} = \frac{n-1}{n}\sum_{i=1}^n{\norm{\phii{\phi^k_{i}}}^2} + \Eik{\norm{\phik{\phi^{k+1}_{i_k}}}^2}.
\]

We can thus arrange the terms as
\[
\begin{aligned}
	&\left(\frac{n}{\alpha^2} + \frac{2(\alpha + \mu)n}{\alpha^2(L - \mu)}\right)\Eik{\frac{1}{n}\sum_{i=1}^n{\norm{\phii{\phi^{k+1}_{i}}}^2}} + \left(1 + \frac{\mu}{\alpha}\right)^2 \Eik{\norm{x_{k+1} - \xs}^2}\\
	\leq{}& \left(\frac{n}{\alpha^2} + \frac{2(\alpha + \mu)(n-1)}{\alpha^2(L - \mu)}\right)\cdot\frac{1}{n}\sum_{i=1}^n{\norm{\phii{\phi^k_{i}}}^2} +  \norm{x_k - \xs}^2.
\end{aligned}
\]

In view of the Lyapunov function, we choose $\alpha$ to be the positive root of the following equation:
\[
\left(1 + \frac{\mu}{\alpha}\right)^2  \left(1 - \frac{2(\alpha + \mu)}{n(L - \mu) + 2n(\alpha + \mu)}\right) = 1.
\]

Let $q = \frac{\alpha}{\mu} > 0$, the above is a cubic equation:
\[
s(q) \triangleq 2q^3 - (4n-6)q^2 - (2n\kappa + 4n - 6) q - (n\kappa + n - 2) = 0,
\]
which has a unique positive root (denoted as $q^{\star}$).

Note that $s(-\infty) < 0, s(-\frac{1}{2}) = \frac{1}{4}$ and $s(0) \leq 0$. These facts suggest that if for some $u > 0$, $s(u) > 0$, we have $q^{\star} < u$. It can be verified that $s(2n + \sqrt{n\kappa}) > 0$, and thus $q^\star= O(n + \sqrt{n\kappa})$.

\section{Experimental setup}
\label{app:exp}

We ran experiments on an HP Z440 machine with a single Intel Xeon
E5-1630v4 with 3.70GHz cores, 16GB RAM, Ubuntu 18.04 LTS with GCC 4.8.0, MATLAB R2017b. We were optimizing the following binary problems with $a_i \in \R^d$, $b_i \in \lbrace -1, +1\rbrace$, $i\in[n]$:
\[
\begin{split}
	\ell_2\textup{-Logistic Regression: }& \frac{1}{n} \sum_{i=1}^n {\log{\big(1 + \exp{(-b_i\innr{a_i,x})}\big)}} + \frac{\mu}{2} \norm{x}^2,\\
	\textup{Ridge Regression: }& \frac{1}{2n} \sum_{i=1}^n {(\innr{a_i,x} - b_i)^2} + \frac{\mu}{2} \norm{x}^2.
\end{split}
\]

We used datasets from the LIBSVM website \citep{LIBSVM}, including \textsf{a9a} (32,561 samples, 123 features), \textsf{covtype.binary} (581,012 samples, 54 features), \textsf{w8a} (49,749 samples, 300 features), \textsf{ijcnn1} (49,990 samples, 22 features). We added one dimension as bias to all the datasets.

We choose SAGA and Katyusha as the baselines in the finite-sum experiments due to the following reasons: SAGA has low iteration cost and good empirical performance with support for non-smooth regularizers, and is thus implemented in machine learning libraries such as scikit-learn \citep{scikit-learn}; Katyusha achieves the state-of-the-art performance for ill-conditioned problems\footnote{\citet{kw:SSNM} shows that SSNM can be faster than Katyusha in some cases. In theory, SSNM and Katyusha achieve the same rate if we set $m=n$ for Katyusha (both require $2$ oracle calls per-iteration). In practice, if $m=n$, they have similar performance (SSNM is often faster). Considering the stability and memory requirement, Katyusha still achieves the state-of-the-art performance both theoretically and empirically.}.

\section{Analyzing NAG using Lyapunov function}
\label{app-NAG}
In this section, we review the convergence of NAG in the strongly convex setting for a better comparison with the convergence guarantee and proof of G-TM. This Lyapunov analysis has been similarly presented in many existing works, e.g., \citep{wilson2016lyapunov, pmlr-v70-hu17a, potential, paquette2019potential}. We adopt a simplified version of NAG in Algorithm \ref{Uni_Acc} ($1$-memory accelerated methods, \citep{u_AGD}) and only consider constant parameter choices. It is known that NAG can be analyzed based on the following Lyapunov function ($\lambda > 0$):
\begin{equation}\label{NAG-Lya}
	T_k = f(x_k) - f(\xs)\ + \frac{\lambda}{2} \norm{z_k - \xs}^2,
\end{equation}
which is somehow suggested in the construction of the \textit{estimate sequence} in \citet{AGD3}. This choice requires neither $f(x_k) - f(\xs)$ nor $\norm{z_k - \xs}^2$ to be monotone decreasing over iterations, which is called the non-relaxational property in \citet{AGD1}. By re-organizing the proof in \citet{AGD3} under the notion of Lyapunov function, we obtain the per-iteration contraction of NAG in Theorem \ref{NAG-contract}.
\begin{algorithm}[t]
	\caption{Nesterov's Accelerated Gradient (NAG)}
	\label{Uni_Acc}
	\renewcommand{\algorithmicrequire}{\textbf{Input:}}
	\renewcommand{\algorithmicensure}{\textbf{Output:}}
	\begin{algorithmic}[1]
		\REQUIRE Parameters $\alpha > 0, \tau_y, \tau_x \in ]0, 1[$ and initial guesses $x_0, z_0 \in \R^d$, iteration number $K$.
		\FOR{$k=0,\ldots, K-1$} 
		\STATE $y_{k} = \tau_{y} z_k + \left(1 - \tau_{y}\right) x_k$.
		\STATE $z_{k+1} = \arg\min_{x} \Big\{\innr{\pf{y_k}, x} + (\alpha/2) \norm{x - z_k}^2 + (\mu/2)	\norm{x - y_k}^2 \Big\}$.
		\STATE $x_{k+1} = \tau_{x} z_{k+1} + (1 - \tau_{x}) x_k$.
		\ENDFOR
		\ENSURE $x_K$.
	\end{algorithmic}
\end{algorithm}

\begin{app-theorems} \label{NAG-contract}
	In Algorithm~\ref{Uni_Acc}, suppose we choose $\alpha, \tau_x, \tau_y$ under the constraints \eqref{NAG-constraint}, the iterations satisfy the contraction \eqref{NAG-per-iteration} for the Lyapunov function \eqref{NAG-Lya}.\\
	\begin{minipage}{0.45\linewidth}
		\rule{0pt}{1pt}\begin{equation}\label{NAG-constraint}
			\begin{cases}
				\alpha \geq \frac{L(1 - \tau_x)\tau_y}{1 - \tau_y}, \tau_x \geq \tau_y,  \\
				\mu \geq \frac{L(\tau_x - \tau_y)}{1 - \tau_y},  \\
				\left(1 + \frac{\mu}{\alpha}\right) (1 - \tau_x) \leq 1.
			\end{cases}
		\end{equation}
	\end{minipage}\qquad
	\begin{minipage}{0.47\linewidth}
		\begin{equation}\label{NAG-per-iteration}
			\begin{aligned}
				&\text{With } \lambda = (\alpha + \mu)\tau_x, \\
				&T_{k+1} \leq \left(1 + \frac{\mu}{\alpha}\right)^{-1} T_k, \text{ for }k\geq 0.
			\end{aligned}
		\end{equation}
	\end{minipage}
\end{app-theorems}

When the inequalities in constraints \eqref{NAG-constraint} (except $\tau_x \geq \tau_y$) hold as equality, we derive the standard choice of NAG: $\alpha = \sqrt{L\mu} - \mu, \tau_y = (\sqrt{\kappa} + 1)^{-1}, \tau_x = (\sqrt{\kappa})^{-1}$. By substituting this choice and eliminating sequence $\{z_k\}$, we recover the widely-used scheme (Constant Step scheme III in \citet{AGD3}):
\[
\begin{aligned}
	x_{k+1} &= y_k - \frac{1}{L} \pf{y_k},\\ 
	y_{k+1} &= x_{k+1} + \frac{\sqrt{\kappa} - 1}{\sqrt{\kappa} + 1} (x_{k+1} - x_k).
\end{aligned}
\]
Telescoping \eqref{NAG-per-iteration}, we obtain the original guarantee of NAG (cf. Theorem 2.2.3 in \citet{AGD3}),
\[
f(x_{K}) - f(\xs) + \frac{\mu}{2}\norm{z_{K} - \xs}^2 \leq{} \left(1 - \frac{1}{\sqrt{\kappa}}\right)^K\left(f(x_0) - f(\xs) + \frac{\mu}{2}\norm{z_0 - \xs}^2\right).
\]
If we regard the constraints~\eqref{NAG-constraint} as an optimization problem with a target of minimizing the rate factor $(1 + \frac{\mu}{\alpha})^{-1}$, the rate factor $1 - 1/\sqrt{\kappa}$ is optimal. Combining $\alpha \geq \frac{L(1 - \tau_x)\tau_y}{1 - \tau_y}$ and $\mu \geq \frac{L(\tau_x - \tau_y)}{1 - \tau_y}$, we have $\alpha \geq L\tau_x - \mu$. To minimize $\alpha$, we fix $\alpha = L\tau_x - \mu$, and it can be easily verified that in this case, the smallest rate factor is achieved when $\left(1 + \frac{\mu}{\alpha}\right) (1 - \tau_x) = 1$. Note that these arguments do not consider variable-parameter choices and are limited to the current analysis framework only.

Denote the initial constant as $C^{\text{NAG}}_0 \triangleq f(x_0) - f(\xs) + \frac{\mu}{2}\norm{z_0 - \xs}^2$. This guarantee shows that in terms of reducing $\norm{x - \xs}^2$ to $\epsilon$, sequences $\{x_k\}$ and $\{z_k\}$
have the same iteration complexity $\sqrt{\kappa} \log{\frac{2C_0^{\text{NAG}}}{\mu\epsilon}}$. Since $\{y_k\}$ is a convex combination of them, it also converges with the same complexity.

\subsection{Proof of Theorem \ref{NAG-contract}}
For the convex combination $y_k = \tau_{y} z_k + \left(1 - \tau_{y}\right) x_k$, we can use the trick in Lemma \ref{func-contract} to obtain
\begin{align}
	f(y_k) - f(\xs) \leq{}& \frac{1 - \tau_y}{\tau_y}\innr{\pf{y_k}, x_k - y_k} + \innr{\pf{y_k}, z_k - \xs} - \frac{\mu}{2} \norm{y_k - \xs}^2 \nonumber\\
	={}& \frac{1 - \tau_y}{\tau_y}\innr{\pf{y_k}, x_k - y_k}  + \underbrace{\innr{\pf{y_k}, z_k - z_{k+1}}}_{R_1} \label{P1}\\
	&+ \underbrace{\innr{\pf{y_k}, z_{k+1} - \xs}}_{R_2}- \frac{\mu}{2} \norm{y_k - \xs}^2.  \nonumber
\end{align}

For $R_1$, based on the $L$-smoothness, we have
\[
f(x_{k+1}) - f(y_k) + \innr{\pf{y_k}, y_k - x_{k+1}} \leq \frac{L}{2} \norm{x_{k+1} - y_k}^2.
\]

Note that $y_k - x_{k+1} = \tau_x (z_k -  z_{k+1}) + (\tau_y - \tau_x)(z_k - x_k)$, we can arrange the above inequality as
\begin{gather}
	f(x_{k+1}) - f(y_k) + \innr{\pf{y_k}, \tau_x (z_k -  z_{k+1}) + (\tau_y - \tau_x)(z_k - x_k)} \leq \frac{L}{2} \norm{x_{k+1} - y_k}^2, \nonumber\\
	R_1 \leq \frac{L}{2\tau_x} \norm{x_{k+1} - y_k}^2 + \frac{1}{\tau_x} \big(f(y_k) - f(x_{k+1})\big) - \frac{\tau_y - \tau_x}{\tau_x}\innr{\pf{y_k}, z_{k} - x_k}. \label{P2}
\end{gather}

For $R_2$, based on the optimality condition of the $3$rd step in Algorithm~\ref{Uni_Acc}, which is for any $u \in \R^d$,
\[
\begin{aligned}
	\innr{\pf{y_k}  + \alpha (z_{k+1} - z_k) + \mu (z_{k+1} - y_k), u - z_{k+1}} = 0,
\end{aligned}
\]
we have (by choosing $u = \xs$),
\begin{align}
	R_2 ={}& \alpha\innr{z_{k+1} - z_k, \xs - z_{k+1}} + \mu \innr{z_{k+1} - y_k, \xs - z_{k+1}}\nonumber \\
	={}& \frac{\alpha}{2}(\norm{z_k - \xs}^2 - \norm{z_{k+1} - \xs}^2 - \norm{z_{k+1} - z_k}^2) \label{P3}\\
	&+ \frac{\mu}{2} (\norm{y_k - \xs}^2 - \norm{z_{k+1} - \xs}^2 - \norm{z_{k+1} - y_k}^2).\nonumber
\end{align}

By upper bounding \eqref{P1} using \eqref{P2}, \eqref{P3}, we can conclude that
\[
\begin{aligned}
	f(y_k) - f(\xs) \leq{}& \frac{1 - \tau_x}{\tau_x}\innr{\pf{y_k},x_k - y_k} + \frac{1}{\tau_x} \big(f(y_k) - f(x_{k+1})\big) \\
	&+ \frac{\alpha}{2}\left(\norm{z_k - \xs}^2 - \left(1 + \frac{\mu}{\alpha}\right)\norm{z_{k+1} - \xs}^2\right) \\ 
	&+ \frac{L}{2\tau_x} \norm{x_{k+1} - y_k}^2 - \frac{\alpha}{2}\norm{z_{k+1} - z_k}^2- \frac{\mu}{2}\norm{z_{k+1} - y_k}^2,
\end{aligned}
\]

Re-arrange the terms,
\begin{equation}\label{P4}
	\begin{aligned}
		f(x_{k+1}) - f(\xs) \leq{}& (1 - \tau_x)\big(f(x_k) - f(\xs)\big) + \frac{\alpha\tau_x}{2}\left(\norm{z_k - \xs}^2 - \left(1 + \frac{\mu}{\alpha}\right)\norm{z_{k+1} - \xs}^2\right) \\ 
		&+ \frac{L}{2} \norm{x_{k+1} - y_k}^2 - \frac{\alpha\tau_x}{2}\norm{z_{k+1} - z_k}^2- \frac{\mu\tau_x}{2}\norm{z_{k+1} - y_k}^2.
	\end{aligned}
\end{equation}

Note that the following relation holds:
\[
x_{k+1} - y_k = \tau_x \left(\frac{(1 - \tau_x)\tau_y}{(1 - \tau_y)\tau_x} (z_{k+1} - z_k) + \frac{\tau_x - \tau_y}{(1 - \tau_y)\tau_x} (z_{k+1} - y_k)\right),
\] 
and thus if $\tau_x \geq \tau_y$, based on the convexity of $\norm{\cdot}^2$, we have
\[
\frac{L}{2}\norm{x_{k+1} - y_k}^2 \leq \frac{L(1 - \tau_x)\tau_x\tau_y}{2(1 - \tau_y)} \norm{z_{k+1} - z_k}^2 + \frac{L(\tau_x - \tau_y)\tau_x}{2(1 - \tau_y)}\norm{z_{k+1} - y_k}^2.
\] 

Finally, suppose that the following relations hold
\[
\begin{cases}
	\tau_x \geq \tau_y, \\
	\alpha \geq \frac{L(1 - \tau_x)\tau_y}{1 - \tau_y},  \\
	\mu \geq \frac{L(\tau_x - \tau_y)}{1 - \tau_y},  \\
	\left(1 + \frac{\mu}{\alpha}\right) (1 - \tau_x) \leq 1,
\end{cases}
\]
we can arrange \eqref{P4} as
\[
\begin{aligned}
	&f(x_{k+1}) - f(\xs) + \frac{\alpha\tau_x}{2}\left(1 + \frac{\mu}{\alpha}\right)\norm{z_{k+1} - \xs}^2\\ \leq{}& \left(1 + \frac{\mu}{\alpha}\right)^{-1}\left(f(x_k) - f(\xs) + \frac{\alpha\tau_x}{2}\left(1 + \frac{\mu}{\alpha}\right)\norm{z_k - \xs}^2\right),
\end{aligned}
\]
which completes the proof.
\end{document}